\documentclass[manuscript,screen]{acmart}
\AtBeginDocument{%
	\providecommand\BibTeX{{%
			\normalfont B\kern-0.5em{\scshape i\kern-0.25em b}\kern-0.8em\TeX}}}

\setcopyright{acmcopyright}
\copyrightyear{2018}
\acmYear{2018}
\acmDOI{10.1145/1122445.1122456}

\usepackage{graphicx}
\usepackage{amsmath}
\usepackage{multirow}
\usepackage{kotex}
\usepackage{xcolor}
\usepackage{url}
\usepackage{tabularx}
\usepackage{subcaption}
\usepackage{verbatim}
\usepackage{lscape}

\begin{document}
\title{Data Transformation Strategies to Remove Heterogeneity}
\author{Sangbong Yoo}
\email{usangbong@gmail.com}
\orcid{0000-0002-0973-9288}
\affiliation{%
    \institution{Korea Institute of Science and Technology}
    \city{Seoul}
    \country{Republic of Korea}
}

\author{Jaeyoung Lee}
\email{wodud5314@gmail.com}
\orcid{0000-0002-1789-7016}
\affiliation{%
    \institution{Hanyang University}
    \city{Seoul}
    \country{Republic of Korea}
}

\author{Chanyoung Yoon}
\email{vfgtre8746@gmail.com}
\orcid{0000-0002-9784-0238}
\affiliation{%
    \institution{Sejong University}
    \city{Seoul}
    \country{Republic of Korea}
}

\author{Geonyeong Son}
\email{handgunzero2@gmail.com}
\orcid{0000-0003-0454-0736}
\affiliation{%
    \institution{Hanyang University}
    \city{Seoul}
    \country{Republic of Korea}
}

\author{Hyein Hong}
\email{gumdung98@naver.com}
\orcid{0000-0003-1410-1788}
\affiliation{%
    \institution{Sejong University}
    \city{Seoul}
    \country{Republic of Korea}
}

\author{Seongbum Seo}
\email{seo@seongbum.com}
\orcid{0000-0002-9582-1674}
\affiliation{%
    \institution{Sejong University}
    \city{Seoul}
    \country{Republic of Korea}
}

\author{Soobin Yim}
\email{tn12qls@gmail.com}
\orcid{0000-0002-9887-8537}
\affiliation{%
    \institution{Sejong University}
    \city{Seoul}
    \country{Republic of Korea}
}

\author{Chanyoung Jung}
\email{ekzmsltm6560@naver.com}
\orcid{0009-0001-2284-3506}
\affiliation{%
    \institution{Sejong University}
    \city{Seoul}
    \country{Republic of Korea}
}

\author{Jungsoo Park}
\email{walnut712@naver.com}
\orcid{0009-0004-5530-2767}
\affiliation{%
    \institution{Sejong University}
    \city{Seoul}
    \country{Republic of Korea}
}

\author{Misuk Kim}
\authornote{
    The corresponding author
}
\email{misukkim@hanyang.ac.kr}
\orcid{0000-0002-8623-3088}
\affiliation{%
    \institution{Hanyang University}
    \city{Seoul}
    \country{Republic of Korea}
}

\author{Yun Jang}
\email{jangy@sejong.edu}
\orcid{0000-0001-7745-1158}
\affiliation{%
    \institution{Sejong University}
    \city{Seoul}
    \country{Republic of Korea}
}

\renewcommand{\shortauthors}{Yoo et al.}

\begin{abstract}
Data heterogeneity is a prevalent issue, stemming from various conflicting factors, making its utilization complex. This uncertainty, particularly resulting from disparities in data formats, frequently necessitates the involvement of experts to find resolutions. Current methodologies primarily address conflicts related to data structures and schemas, often overlooking the pivotal role played by data transformation. As the utilization of artificial intelligence (AI) continues to expand, there is a growing demand for a more streamlined data preparation process, and data transformation becomes paramount. It customizes training data to enhance AI learning efficiency and adapts input formats to suit diverse AI models. Selecting an appropriate transformation technique is paramount in preserving crucial data details. Despite the widespread integration of AI across various industries, comprehensive reviews concerning contemporary data transformation approaches are scarce. This survey explores the intricacies of data heterogeneity and its underlying sources. It systematically categorizes and presents strategies to address heterogeneity stemming from differences in data formats, shedding light on the inherent challenges associated with each strategy.
\end{abstract}



\maketitle
\section{Introduction}
\label{sec:intro}
In this paper, we conduct a comprehensive survey of data transformation strategies to mitigate heterogeneity issues arising from varying data formats. This survey intends to tackle the constraints encountered when working with heterogeneous data, primarily due to its innate diversity. We elucidate the compelling need for these transformations and emphasize their crucial role in effectively surmounting the hurdles posed by heterogeneous data.

\subsection{Motivation}
\label{subsec:motivation}
Advancements in technology have produced a substantial collection of data featuring a wide range of distinct attributes. These attributes, which can introduce discrepancies or conflicts within the data, are commonly referred to as heterogeneity, giving rise to the concept of heterogeneous data. Heterogeneity poses challenges for effective data utilization, resulting in the subsequent limitations associated with heterogeneous data.
\begin{itemize}
    \item[\textbf{L1}] Heterogeneity leads to confusion due to its various definitions.
    \item[\textbf{L2}] Determining suitable strategies for removing heterogeneity to meet various objectives is challenging.
    \item[\textbf{L3}] No survey technically classifies data transformation strategies, including the latest AI technology.
\end{itemize}
In this study, we explain the rationale behind our survey, focusing on the limitations (L1, L2, and L3). Data transformation is crucial in harnessing heterogeneous data, especially when dealing with format conflicts. However, due to the dearth of comprehensive technical reviews in existing surveys, numerous researchers and developers encounter difficulties adopting the latest data transformation methods. Hence, through this survey, we aim to furnish a technical categorization of data transformation strategies and provide guidance on their utilization based on the data source and target formats.

\subsubsection{Confusion in the Definition of Heterogeneity:}
The abundance of heterogeneous definitions regarding heterogeneity presents a significant challenge in effectively harnessing diverse data (\textbf{L1}). These varying interpretations of heterogeneity complicate our ability to cope, as they introduce differing understandings of what constitutes heterogeneous data. Heterogeneity in data sources encompasses schema conflicts~\cite{kim1991classifying}, data conflicts~\cite{meng2020survey}, format conflicts~\cite{adnan2019limitations}, and domain conflicts~\cite{maree2015addressing}. Various methods for mitigating heterogeneity are employed, tailored to the specific definitions of heterogeneous data. However, selecting an appropriate technique for heterogeneity reduction remains challenging due to the amalgamation of these diverse interpretations (\textbf{L2}). Researchers have extensively explored strategies for addressing data heterogeneity stemming from schema and data conflicts, primarily within database management. Conversely, there is still a notable absence of a comprehensive investigation into strategies for alleviating heterogeneity caused by format conflicts (\textbf{L3})~\cite{batini1984methodology, bleiholder2009data, hendler2014data, hellerstein2008quantitative, rahm2000data, hong2023how}.

\subsubsection{Challenges in Identifying Strategies for Removing Heterogeneity:}
Identifying a practical approach to mitigate heterogeneity and enable the utilization of heterogeneous data across diverse formats within each specific domain poses a challenging task (\textbf{L2}). Since the introduction of ChatGPT by OpenAI in November 2021, there has been an increasing desire for a multitude of human-like artificial intelligence (AI), including entities such as LLaMa (Large Language Model Meta AI), Alpaca, and DALL$\cdot$E2. Furthermore, readily accessible AI solutions include meeting assistants like Naver's CLOVA~\cite{clova} and Firefiles.ai's Firefiles~\cite{firefiles}, English conversation study tools like Speakeasy Labs' Speak~\cite{speak} and ELSA's ELSA Speak~\cite{elsa}, OS utilization assistants like Apple's Siri~\cite{siri} and Microsoft's Cortana~\cite{cortana}, and mental health care tools like Youper~\cite{youper}. This surge in demand is fueled by the need for substantial quantities of high-quality data stored in diverse formats. These models, deep learning-based ones that serve as the foundation for AI, require different input formats. Text models tokenize text data from diverse sources like CommonCrawl~\cite{commoncrawl} dumps, websites, and books to train~\cite{wenzek2020ccnet}. GPT-3 and LLaMa, for example, were trained using approximately 570GB of preprocessed text data, which included around 8 million web pages. LLaMa used an additional 4.7TB of preprocessed text data for learning~\cite{brown2020language, touvron2023llama, touvron2023llama2}. Alpaca, in turn, fine-tuned LLaMa with 52 thousand text data~\cite{taori2023alpaca}. On the other hand, DALL$\cdot$E2, which is an image generation model, employs training data consisting of approximately 650 million image-text pairs~\cite{ramesh2022hierarchical}. These models necessitate transforming data into different input formats depending on whether they are text-based or image-based. AI platforms such as CLOVA, Firefiles, Speak, ELSA Speak, Siri, Cortana, and Youper assist users by processing and learning from various data formats, including audio, video, tables, and text. The diverse input requirements of deep learning models emphasize the importance of identifying appropriate data transformation strategies to manage the inherent data heterogeneity effectively.

\subsubsection{The insufficiency of surveys focused on data transformation strategies:}
While the latest technologies, including AI, have found extensive applications in various industrial domains, a comprehensive technical framework remains absent for organizing data transformation technologies~(\textbf{L3}). AI is deployed across diverse fields, often as a pre-trained model. Nevertheless, the data preparation process for developing these pre-trained models demands considerable human effort. Unfortunately, data transformation can sometimes result in the loss of specific information~\cite{malik2016big}, posing challenges in identifying appropriate techniques to retain the necessary data details. One available approach to streamline data preparation and reduce human intervention involves employing deep learning techniques for data transformation. Deep learning also presents a solution to mitigate the potential loss of information~\cite{malik2016big} that may occur during the data format transformation process. Data in various formats inherently contains valuable information, encompassing relationships, strengths, representations, and structural elements. Data-driven AI models excel in data transformation by preserving essential information tailored to specific objectives. Various deep-learning studies have already proposed methods for effectively utilizing data stored in diverse formats by addressing heterogeneity. These techniques encompass multimodal learning~\cite{baltrusaitis2019multimodal}, integrating image and text~\cite{karpathy2015deep}, combining graph and textcite{yin2017local}, merging video and text~\cite{venugopalan2015sequence}, assimilating database content with textual information~\cite{dong2013big}, and blending image, text, and graph data~\cite{wang2019dynamic}. Despite the involvement of numerous AI technologies in data transformation, a comprehensive survey encompassing the latest advancements is still lacking~(\textbf{L3}).

\subsection{Survey Goal}
This survey offers a comprehensive overview of strategies for transforming heterogeneous data stored in various formats. Previous research used the term "heterogeneity" with varying definitions, leading to confusion. Our survey provides a clear framework for studying heterogeneous data transformation by offering precise definitions and corresponding solutions based on factors contributing to conflicts. Additionally, we shed light on the practical application of heterogeneous data by summarizing data transformation techniques tailored to popular data formats such as tables, text, images, videos, and graphs. Furthermore, we provide an in-depth exploration of the limitations and challenges associated with heterogeneous data transformation. This survey addresses the limitations of heterogeneous data introduced in Section~\ref{subsec:motivation}.
\begin{itemize}
    \item In order to clarify semantic ambiguities, we introduce a classification of heterogeneity rooted in factors causing data conflicts, along with definitions of heterogeneous data and their respective solutions.
    \item Based on their technical similarity, we categorize data transformation strategies for removing heterogeneity caused by format conflicts in heterogeneous data.
    \item We provide an overview of the challenges arising from the features of data heterogeneity and elucidate relevant data transformation techniques aligned with the intended utilization objectives.
\end{itemize}
The structure of the paper is as follows: In Section~\ref{sec:heterogeneity}, we provide definitions of heterogeneity, strategies for its removal, and relevant surveys as references, categorized by the factors causing conflicts between data. Additionally, we explore data transformation strategies based on the target data format, such as text~(Section~\ref{sec:d2t}) and graph~(Section~\ref{sec:d2g}). Within each section, we also examine the challenges associated with these strategies. To conclude the survey, we summarize our findings and propose future directions for data transformation strategies.
\section{Heterogeneity of Heterogeneous Data}
\label{sec:heterogeneity}
\begin{figure*}
    \centering
    \includegraphics[width=0.8\textwidth]{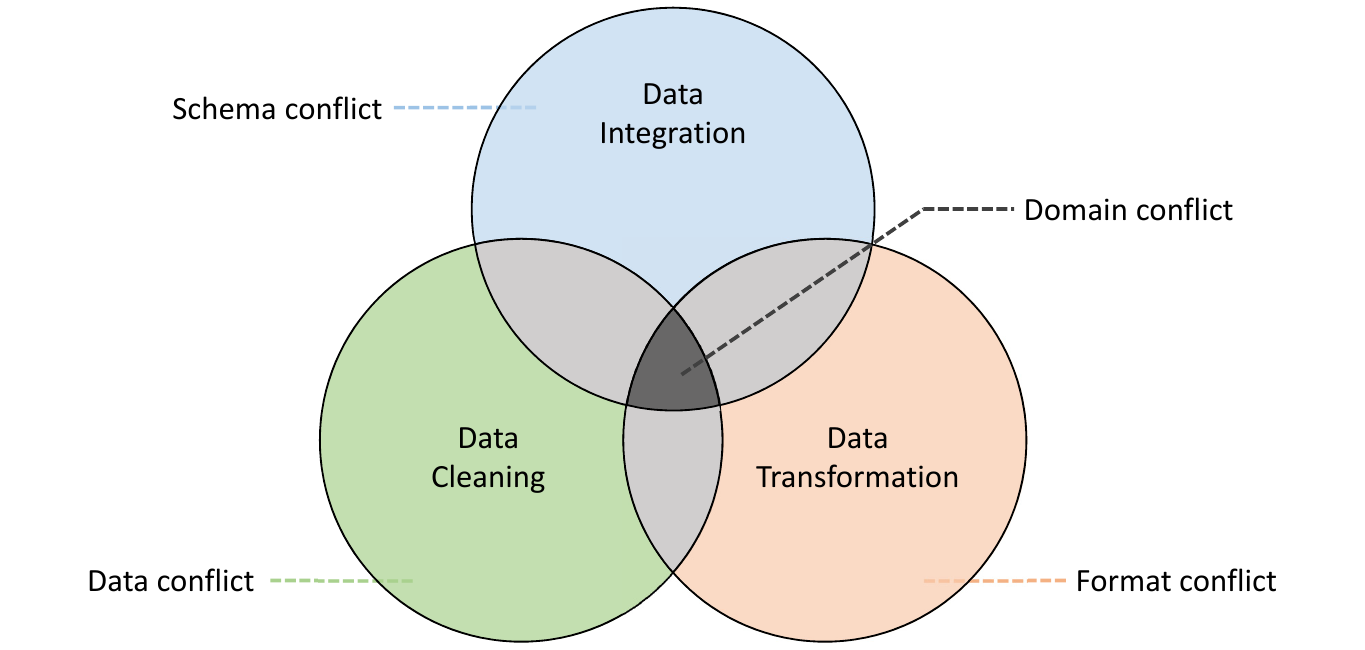}
    \caption{\textcolor{black}{Strategies to remove data heterogeneity according to conflict factors}}
    \label{F:int_heterogeneity}
\end{figure*}
Heterogeneity is an essential feature of heterogeneous data, primarily stemming from disparities in data collection environments and the diversity inherent to various domains~\cite{wang2017heterogeneous}. The heterogeneity within heterogeneous data arises from a multitude of contributing factors. Furthermore, the choice of techniques for mitigating this heterogeneity varies depending on these factors and the intended usage of the data. Figure~\ref{F:int_heterogeneity} illustrates the factors contributing to heterogeneity emergence. Heterogeneity can be traced back to its root causes, namely schema conflict, data conflict, and format conflict. Heterogeneity resulting from domain conflict shows intricate interdependencies with schema, data, and format conflicts. Moreover, techniques for addressing heterogeneity can be broadly categorized into data integration, type conversion, data cleaning, and data transformation. Some conflicts, as illustrated by the overlapping gray area in Figure~\ref{F:int_heterogeneity}, may share common approaches for mitigating heterogeneity. Notably, diverse techniques can be employed to address heterogeneity in the context of domain conflict where all domains intersect, acknowledging that challenges arising from different conflicts magnify the overall complexity of eliminating heterogeneity. In this section, we provide thorough expositions of schema, data, and format conflicts, excluding the specific scenario of domain conflict. For instance, advanced methods such as filtering, translation, modulation, dilation, and downsampling are employed for processing signal data on graphs across diverse domains~\cite{shuman2013emerging}. The unique semantics of numeric values based on the domain can be harnessed for tasks like outlier correction or exception handling through data cleaning and visual analytics~\cite{hong2023how}. Furthermore, techniques that enhance applicability by facilitating knowledge transfer across domains have been introduced~\cite{khetani2023cross}. Consequently, the structure, interpretation, and data analysis practices vary according to the specific domain and use case. Therefore, we explore schema, data, and format conflicts, excluding the specific case of domain conflict in this section.

\subsection{Schema Conflict}
Heterogeneity is an essential feature of heterogeneous data, primarily stemming from disparities in data collection environments and the diversity inherent to various domains~\cite{wang2017heterogeneous}. The heterogeneity within heterogeneous data arises from a multitude of contributing factors. Figure~\ref{F:int_heterogeneity} illustrates the factors contributing to heterogeneity emergence. Furthermore, the choice of techniques for mitigating this heterogeneity varies depending on these factors and the intended usage of the data. Heterogeneity can be traced back to its root causes, namely schema conflict, data conflict, and format conflict. Heterogeneity resulting from domain conflict shows intricate interdependencies with schema, data, and format conflicts. Moreover, techniques for addressing heterogeneity can be broadly categorized into data integration, type conversion, data cleaning, and data transformation. Some conflicts, as illustrated by the overlapping gray area in Figure~\ref{F:int_heterogeneity}, may share common approaches for mitigating heterogeneity. Notably, diverse techniques can be employed to address heterogeneity in the context of domain conflict where all domains intersect, assuming that challenges arising from different conflicts magnify the overall complexity of eliminating heterogeneity.

Data integration, also known as data fusion, serves as a technique to remove heterogeneity from schema conflict. This technology involves fusing heterogeneous data collected from multiple databases or sources~\cite{bleiholder2009data, hendler2014data}. The process of data integration entails schema mapping and duplicate detection~\cite{naumann2006data}. Schema mapping aims to convert the input data schema into the target data schema~\cite{madhavan2001generic}. Duplicate detection focuses on identifying data that co-occurs implications in both source and target data but is expressed inconsistently compared to other attributes~\cite{elmagarmid2007duplicate}. For multi-dimensional data, dimensionality reduction techniques are applied to cluster similar attributes~\cite{jolliffe2002principal, van2008visualizing}. Data integration aims to increase completeness and conciseness, enhancing data usability~\cite{halevy2006data,lenzerini2002data}. While simple rules can govern each step in data integration, human knowledge is essential to creating rules that optimize data usability. Comprehensive descriptions of data integration and fusion techniques can be found in the works of Batini and Lenzerini~\cite{batini1984methodology}, Bleiholder and Naumann~\cite{bleiholder2009data}, Hendler~\cite{hendler2014data}, and Meng et al.~\cite{meng2020survey}.

\subsection{Data Conflict}
Heterogeneity arising from data conflict is primarily driven by data type and value discrepancies, as highlighted in the study~\cite{meng2020survey}. Data that reveals heterogeneity due to data conflict can be described as data where there is a lack of uniformity in data types or variations in the formats of data values. Data types encompass many encoding methods used to store values, including boolean, integer, float, double, var, varchar, character, string, and text. For instance, when dealing with databases, the choice of data type for entering time data may differ depending on the specified time format. Occasionally, time data is stored as integers measured in microseconds, while in other cases, it is represented as text adhering to a specific time format. Moreover, text-based representations of time formats may exhibit slight differences, as exemplified by variations like \textit{YYYY-MM-DD HH:mm:ss} and \textit{YYYY/MM/DD HH:mm:ss}. Data type determination depends on various environmental factors, including the development language, specific database management system (DBMS), and data transmission conditions.

Measurement methods, scaling variations, erroneous data entries, and distinct representations of identical data can also induce data conflict. The geographic location and the purpose of data collection typically influence the preference for measurement methods. Notable differences arising from measurement methods include disparities between feet and meter notations, Fahrenheit and Celsius scales, and pounds and kilograms. Scaling disparities can be attributed to variations in data sampling frequencies, aggregation techniques, value ranges, and normalization methods. For example, data collected from sources like brain waves and eye-tracking~cite{teplan2002fundamentals, cai2023mbrain} may provide differing frequencies based on the sampling rate of the recording device, which can range from 30 Hz to 1000 Hz or even higher. Additionally, weather data may be collected at various intervals, such as 1 or 5 seconds, depending on the specific data collection device. Erroneous data entries can result from various causes, with one of the simplest examples being assigning default values. Default values for missing or erroneous data may vary and are determined by user-defined rules established during the data collection phase. These defaults could include values like \textit{-999}, \textit{0}, or \textit{NULL}, among others. Furthermore, differences in representing the same data~cite{kim1991classifying} enclose variations in letter case~(e.g., "TOWN" and "town"), distinct string representations for identical information~(e.g., formal address format and abbreviations), and variations in data encoding methods~(e.g., English and numeric representations of grades).

Various techniques are employed to mitigate heterogeneity resulting from data conflict, including type conversion and data cleaning. Type conversion directly addresses data type conflicts by performing explicit transformations of data types. Data cleaning, on the other hand, is a set of techniques aimed at enhancing data quality, and it inherently includes the removal of data conflicts within its processes. Several techniques aid in eliminating data conflicts within the domain of data cleaning. These include processes for handling outliers and missing values, re-sampling, and data normalization. Furthermore, data cleaning techniques encompass preprocessing data from different measurement methods and dealing with diverse representations of the same data. Re-sampling resolves conflicts arising from variations in data collection frequencies. At the same time, data normalization eliminates conflicts by computing weights under consistent conditions, thereby mitigating differences in attribute ranges or data collection environments. Authors such as Hellerstein~\cite{hellerstein2008quantitative}, Rahm and Do~\cite{rahm2000data}, and Hong et al.~\cite{hong2023how} offer comprehensive overviews of methodologies and analytical techniques related to data cleaning.

\subsection{Format Conflict}
Heterogeneity resulting from format conflict arises due to differences in how data is encoded within the storage system~\cite{adnan2019limitations}. Standard data formats that we frequently encounter include table, text, image, video, and graph. Format-based heterogeneity is prevalent because data is often stored in different formats based on the data collection environment and the intended use. For example, in the case of traffic information, even within the same geographical region, there can be variations in how road data is collected, such as using CCTV cameras or capturing traffic volume in tabular form. Similarly, records of meetings can be saved in audio or text formats, and data extracted from web articles through crawling may be stored in table or text formats. Data is stored in various formats based on its intended use and the efficiency of the data collection environment~\cite{malik2016big}. Consequently, heterogeneity resulting from format conflicts is a challenge that we readily encounter and is one of the first issues we address during the data preparation process.

Data transformation is a technique employed to address the heterogeneity of format conflicts. This survey places its primary emphasis on the exploration of these methodologies. With the growing interest in deep learning, a fundamental component of artificial intelligence, data transformation techniques have experienced a notable increase in usage, particularly in the context of preparing training data. Data is predominantly stored in table, text, image, video, and numerical format. Data transformation entails converting data stored in diverse formats into the desired target format to resolve conflicts related to formatting. Data transformation also encompasses adapting data into an input format suitable for practical use. In this survey, we systematically categorize strategies for converting prevalent data formats, including table, text, image, video, and numerical data, into formats that are conducive to both human understanding and alignment with model requirements, specifically text, graph, and visual formats, based on their suitability and applicability.
\section{Data-to-Text Transformation}
\label{sec:d2t}
Text format is one of the most easily understandable data formats, as it effectively conveys structured and unstructured information using natural language. Due to this advantage, data-to-text transformation techniques find diverse applications, including text generation, keyword extraction, topic extraction, document summarization, and the generation of training datasets while minimizing information loss. This section provides a comprehensive overview of techniques for transforming tables, images, videos, and source texts into text format, such as extracted keywords, extracted topics, and summarized text.
\subsection{Table-to-Text Transformation}
\label{subsec:table2text}
\begin{figure*}
    \centering
    \includegraphics[width=0.9\textwidth]{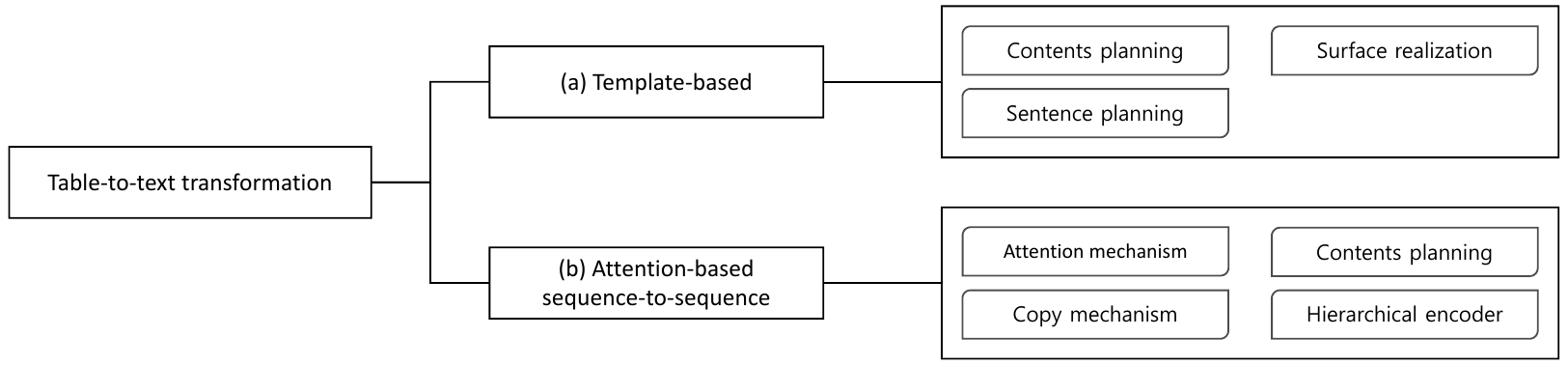}
    \caption{Strategies of table-to-text transformation. Table-to-text is classified into (a) template-based method and (b) attention-based sequence-to-sequence method.}
    \label{F:table_to_text}
\end{figure*}
A table represents structured data, similar to a database, which is easily managed but often challenging for people to comprehend or utilize. Consequently, transforming table data into text is a common practice for its service utilization. Converting table data into text, also known as table-to-text transformation, involves deriving textual information from the table, earning it the designation of text generation. This method is categorized into technical classifications such as template-based and attention-based methods, as illustrated in Figure~\ref{F:table_to_text}, focusing on addressing the limitations of existing text generation techniques.

\subsubsection{Template-based Method}
\label{subsubsec:tab2ttemp}
The template-based method is commonly employed to transform structured data in a table into simple text. This method constructs a pipeline and generates text from the table by modeling structured data from predefined schemas~\cite{liang2009learning, chen2008learning}. For instance, it is employed to create weather forecasts by utilizing structured weather records~\cite{chen2008learning} or in sportscasting by leveraging event records from sports events~\cite{liang2009learning}. 
The pipeline for generating text from a table typically includes content planning, which involves selecting input content and determining the output text structure. It also incorporates sentence planning, which determines sentence structure and lexical content. A surface realization module also converts the defined sentence planning into a coherent string. However, the template-based method has limitations as it may not cover all scenarios beyond the predefined schema. In such cases, it often necessitates the development of new models based on the intended use and the underlying base table dataset.

\subsubsection{Attention-based Sequence-to-Sequence Method}
\label{subsubsec:tab2tattention}
The attention-based sequence-to-sequence method is proposed for generating complex forms of text from tables and for versatile applications. The attention-based sequence-to-sequence model preserves information, such as table fields or sentences, which may get lost in the compression of the encoder as the sequence grows longer~\cite{luong2015effective}. Studies on attention-based sequence-to-sequence models are categorized into altering the encoder-decoder structure or enhancing the text generation pipeline. Unlike the template-based approach, this method requires high-quality data and uses an end-to-end trained neural network model to transform the complex structure of a table into text, not confined by predefined schemas. Additionally, this method generates more fluent and faithful text by adopting attention-based sequence-to-sequence models.

Wiseman et al.~\cite{wisemanACL2017challenges} proposed a model adding a copy mechanism for vocabulary-registered words and input source tokens to the decoder, creating natural text similar to the ground truth. Liu et al.~\cite{liuAAAI2018table} introduced a new structure-aware sequence-to-sequence architecture consisting of a field-gating encoder with dual attention and a description generator. Architectures like the two-stage structure~\cite{liu2019hierarchical} focusing on learning precise meanings within the table source or a hierarchical encoder with layers for timeline, rows, and columns~\cite{gongACL2019table} have also been proposed. Furthermore, research has enhanced the text generation pipeline for more natural table-to-text conversions~\cite{shaAAAI2018order, puduppully2019data, gong2020enhancing}. Modules such as a dispatcher considering the order of contents for pipeline enhancement~\cite{shaAAAI2018order}, content selection gate mechanism~\cite{puduppully2019data}, and introducing ranking loss for contextual and numerical value comprehension~\cite{gong2020enhancing} have been explored.

\subsubsection{Dataset}
Table-to-text transformation using deep learning models requires a large training dataset. Table~\ref{table:tab2text_dataset} presents datasets utilized in previous studies to improve model capabilities. The datasets have been employed to train and evaluate deep learning models for generating text from tabular data.
\begin{table}[!htbp]
    \setlength\tabcolsep{0pt}
    \setlength\thickmuskip{0mu}
    \setlength\medmuskip{0mu}
    \small
    \caption{\textcolor{black}{Datasets used for the table-to-text generation}}
    \centering
    \label{table:tab2text_dataset}
    \begin{tabularx}{\textwidth}{p{0.15\textwidth}p{0.51\textwidth}p{0.13\textwidth}p{0.11\textwidth}p{0.05\textwidth}p{0.05\textwidth}}
        \toprule
            \textbf{Dataset} & \textbf{Description} & \centering \textbf{Open} & \textbf{Venue} & \textbf{Year} & \textbf{Ref} \\
        \midrule
        WikiBio & A biography dataset consisting of Wikipedia info boxes and text & \centering O & EMNLP & 2016 & \cite{lebret2016neural} \\
        \hline
        RotoWire & A dataset composed of tables containing NBA statistical data and game summaries & \centering O & EMNLP & 2017 & \cite{wisemanACL2017challenges} \\
        \hline
        MLB & A dataset composed of tables containing MLB statistical data and game summaries & \centering O & ACL & 2019 & \cite{puduppully2019data} \\
        \hline
        WikiTableText & A dataset composed of table data and text to express a specific table area & \centering O & AAAI & 2018 & \cite{baoAAAI2018table} \\
        \hline
        \begin{tabular}[c]{@{}l@{}}WikiPerson\\ \& Animal\end{tabular} & Datasets containing structured KBs and natural language descriptions of human and animal entities based on Wikipedia dumps and Wikidata & \centering O & ICNLG & 2018 & \cite{wang2018describing} \\
        \hline
        ToTTo & An open-domain dataset created using a new Annotation process, with a controlled text generation task that can be used to evaluate model hallucinations & \centering O & EMNLP & 2020 & \cite{parikh2020totto} \\
        \hline
        numericNLG & A dataset containing numerical tables and descriptive texts from science papers & \centering O & ACL & 2021 & \cite{suadaa2021towards} \\
    \bottomrule
    \end{tabularx}
\end{table}
\subsubsection{Discussion and Challenges}
\label{dis:table2text}
The template-based and attention-based sequence-to-sequence methods presented in Section~\ref{subsubsec:tab2ttemp} and Section~\ref{subsubsec:tab2tattention} respectively focus on generating natural and fluent text. However, existing models often struggle to describe the regions or convey knowledge within the data due to a lack of reflection of the table characteristics. Moreover, text generation has limitations such as minimal resources, handling different languages, overly complex models, and maintaining fidelity to the table source.
Consequently, ongoing research aims to address these limitations posed by prior studies. 

Bao et al.~\cite{baoAAAI2018table} propose a specially tailored model to describe table regions, aiming to overcome the challenge of overlooked table areas that contain semantic information. Wang et al.~\cite{wang2018describing} present a model that can effectively describe a knowledge base table using a pointer network. To address challenges such as text generation with limited resources~\cite{ma2019key, wang2021sketch}, transformer-based text generation~\cite{wang2020towards}, utilization of large-scale pre-trained language models~\cite{narayan2020stepwise, gong2020tablegpt}, text generation based on numerical reasoning~\cite{suadaa2021towards}, and reinforcement learning with large-scale corpus prototype memory~\cite{su2021few} have been proposed. Another approach to tackle the issue of hallucination, where generated text diverges from the table source, involves aligning the source table and reference text to produce faithful text~\cite{liu2021towards}. A multi-branch architecture for granular access at the word level has also been suggested~\cite{rebuffel2022controlling}. In summary, various research initiatives are underway to refine existing models, address limitations, and enhance the quality and accuracy of text generation from tables, thus advancing the capabilities and applications of these techniques.

Advancements in neural network models have led to increased fluency in table-to-text transformations. However, as heterogeneous data finds broader applications across industries, achieving more flexible text generation remains challenging. The issue of hallucination is highlighted as a significant challenge, causing the loss of crucial source information due to text not remaining faithful to the table source~\cite{liu2021towards}. Moreover, existing text generation models target only single tables, and the available training data is structured solely for generating text from individual tables. Consequently, generating text for multiple tables or tables with different structures remains difficult, reducing the versatility of table-to-text transformation models. Additionally, numeric values within table data are recognized as vital information. However, due to insufficient research in converting numeric sources into understandable text, a considerable disparity remains from text generated by humans. Hence, research on understanding numeric data is a critical challenge that needs addressing to broaden the utilization of table-to-text transformation applications~\cite{suadaa2021towards}. 
\subsection{Text-to-Text Transformation}
\begin{figure*}
    \centering
    \includegraphics[width=.99\textwidth]{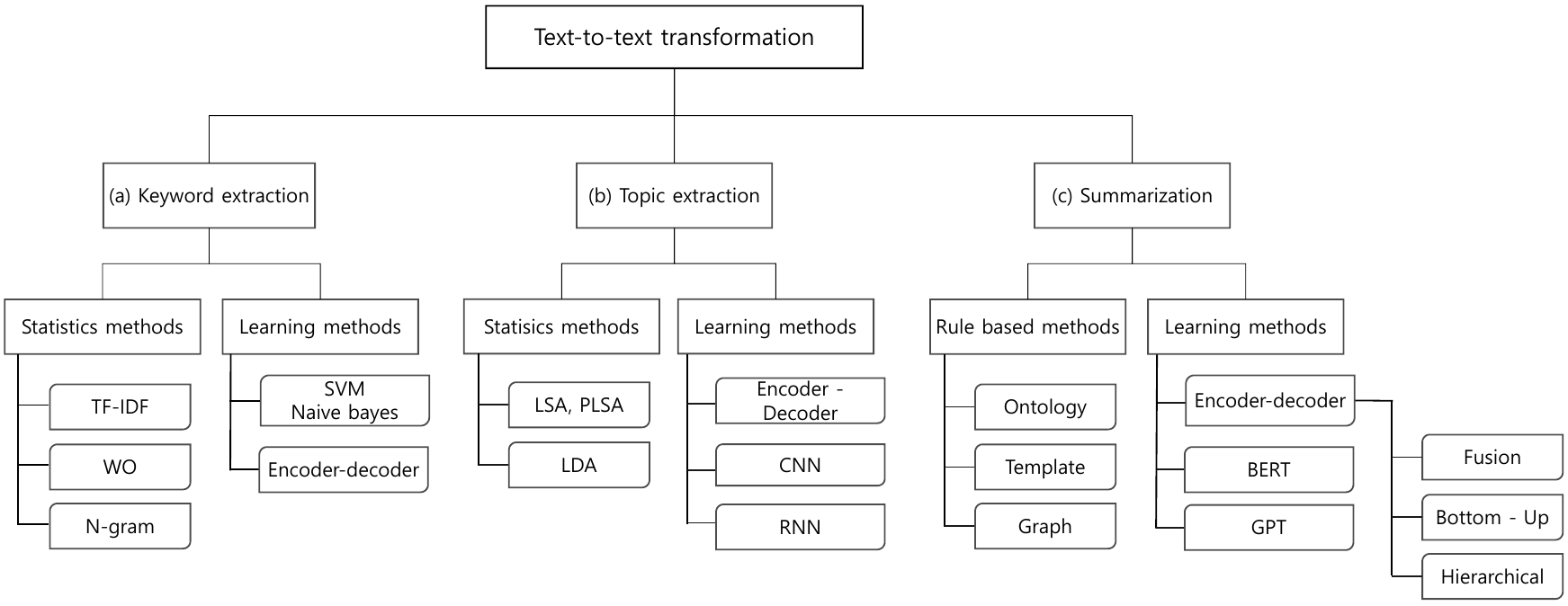}
    \caption{Strategies of text-to-text. (a) is keyword extraction, (b) is topic extraction, and (c) is summarization.}
    \label{F:text_to_text}
\end{figure*}
Text-to-text transformation is a task that involves generating text in a different format while preserving the meaning of the original text data. This versatile approach finds applications in various tasks such as translation, summarization, dialogue systems, and question-answering. Depending on the specific purpose, text-to-text methods utilize different datasets and models. This section focuses on the text-to-text transformation techniques employed for keyword extraction, topic extraction, and summarization tasks. These techniques are crucial in efficiently transforming text data into more concise and relevant forms. Figure~\ref{F:text_to_text} shows text-to-text transformation strategies.

\subsubsection{Keyword Extraction Method}
Keywords serve as representative words or phrases that capture the essence of text data. The keyword extraction technique is employed to identify crucial information within documents. Statistical and learning methods are commonly used to achieve this, as shown in Figure~\ref{F:text_to_text} (a). The statistical keyword extraction technique identifies keywords based on word frequency~\cite{ramos2003using} and word distribution~\cite{wartena2010keyword, matsuo2004keyword}. On the other hand, learning methods~\cite{armouty2019automated, sakakibara1993text, yih2006finding, wu2005domain, mou-etal-2016-sequence, li2018guiding, ge2021keywords} have been proposed to perform more targeted and specific keyword extraction from documents.

The concept of statistical keyword extraction techniques was initially proposed by Luhn et al.~\cite{luhn1957statistical}. These techniques have gained popularity due to their simplicity in calculations and not requiring vast amounts of data. Several statistical methods are used for keyword extraction, including Term Frequency-Inverse Document Frequency~(TF-IDF), Word Occurrence~(WO), and N-gram. TF-IDF identifies keywords by considering the frequency of a word appearing in a single sentence and its frequency across multiple sentences~\cite{luhn1957statistical, ramos2003using}. Although it involves straightforward calculations, TF-IDF treats synonyms as distinct words and does not consider sentence order and semantic relationships~\cite{wartena2010keyword}. WO is a technique that defines the co-occurrence distribution of words. It extracts keywords by comparing the defined word distribution with the Kullback-Leibler~(KL) divergence of the text data distribution~\cite{wartena2010keyword, matsuo2004keyword}. While TF-IDF and WO express keywords as single words, N-gram represents keywords as sequences of words of arbitrary length~\cite{campos2018text, hulth2003improved}. N-gram considers the sequence of sentences by specifying the sequence length $N$, allowing it to extract semantic keywords. However, N-grams have limitations, as their performance varies depending on the specified $N$ and may overlook sparsity issues.

Learning-based keyword extraction models outperform statistical techniques in learning the structure of documents by leveraging labeled data, allowing them to extract semantic keywords that consider the meaning of documents~\cite{sakakibara1993text, yih2006finding, wu2005domain}. Machine learning models used for keyword extraction include Support Vector Machine~(SVM) and Naive Bayes. SVM learns statistical characteristics of word frequency calculations extracted from labeled data using TF-IDF, and the extracted statistical features are used to select candidates for keyword extraction. However, keyword extraction using candidate selection tends to rely heavily on statistical characteristics~\cite{armouty2019automated}. The technique using Naive Bayes also extracts keywords using a score measurement method based on TF-IDF, but it assumes that keyword features are independent and follow a normal distribution. Deep learning models, such as sequence-to-sequence, Long Short-Term Memory~(LSTM), and attention mechanism-based models, are also employed for keyword extraction. For instance, Mou et al.~\cite{mou-etal-2016-sequence} re-extract semantic keywords using pointwise mutual information in a sequence-to-sequence model. The bi-LSTM model-based technique considers sentence order and meaning in both directions to re-extract keywords~\cite{duan2021oilog, wang2017keyword}. In the encoder-decoder framework, a method of extracting keywords by incorporating an attention mechanism during the context vector update process was proposed~\cite{li2018guiding}. Learning model-based keyword extraction techniques yield more accurate results than statistical methods and traditional machine learning. However, they require more time for model training, and there are limitations in interpreting and analyzing the keyword extraction process.

\subsubsection{Topic Extraction Method}
Topic extraction involves identifying sentences or sentence combinations containing semantic content from text data. Topics are extracted using statistical and learning methods, as shown in Figure~\ref{F:text_to_text}~(b). Statistical models for topic extraction include Latent Dirichlet Allocation~(LDA), Latent Semantic Analysis~(LSA), and Probabilistic Latent Semantic Analysis~(PLSA). LSA uses singular value decomposition to extract topics, approximating linear combinations of singular values~\cite{deerwester1990indexing}. PLSA combines LSA and conditional probability~\cite{hofmann1999probabilistic}. There are also hybrid models, such as W2V-LSA, which combines word2vec and LSA to capture the context of a sentence~\cite{kim2020word2vec}. However, LSA and PLSA show poor performance in new documents. LDA was proposed to overcome this. LDA is a widely used method that predicts potential topics based on word distributions found in the text data and known word distributions by topic. It assumes that potential topics are generated based on a probability distribution and predicts them by comparing probability distributions of word sets. To improve LDA performance, additional calculations of correlation and density are conducted~\cite{cao2009density}. Afterward, Menon et al.~\cite{menon2017semantics} proposed a model combining LDA and LSA to understand the meaning of documents better.

To extract topics using deep learning models, sequential models such as Sequence-to-Sequence~\cite{xing2017topic}, RNN~\cite{miao2017discovering, wei2019translating}, and LSTM are used. Some models use pre-trained LDA to obtain latent topics and combine them with a sequence-to-sequence model~\cite{sakakibara1993text}. Others propose models that assign topic parameter distributions to RNNs to discover multiple topics~\cite{miao2017discovering} or identify bilingual topics using the hidden state of encoder and decoder~\cite{wei2019translating}. These deep learning-based approaches improve the performance of topic extraction by better understanding the meaning of documents compared to statistical methods. However, they still have limitations regarding model interpretability and analysis of the topic extraction process.

\subsubsection{Summarization Method}
Summarization is the process of condensing lengthy text data into shorter forms. Two main approaches have been proposed for summarization: rule-based and learning-based methods on the encoder-decoder framework, as shown in Figure~\ref{F:text_to_text}~(c). Rule-based methods are categorized into ontology, template, and graph-based techniques. Okumura et al.~\cite{okumura2015automatic} proposed an ontology-based approach that establishes relationships between sentences in the input text and descriptive labels, such as words or sentences, to generate high-weight sentences for the summary text. They applied the weights suggested by Tseng~\cite{tseng2006toward} to the ontology. Oya et al.~\cite{oya2014template} developed a template technique that employs multi-sentence convergence. They summarized documents using templates and word graphs created with noun phrases extracted from POS tagging and hypernyms. Gerani et al.~\cite{gerani2014abstractive} utilized obtained sentences as templates, considering the discourse structure of the text, and used them for summarization. Furthermore, graph-based summarization techniques, such as LexRank~\cite{erkan2004lexrank} and TextRank~\cite{mihalcea2004textrank}, have been proposed. These methods summarize text data by constructing graphs based on lexical attributes and sentence connections using overlap frequencies.

The encoder-decoder framework comprises encoder and decoder modules that generate succinct text data. The encoder assimilates text data, creating a context vector that compresses document information. Utilizing this context vector, the decoder sequentially produces words related to the given context. Text summarization studies employ sequential models like the Neural Network Language Model~(NNLM)~\cite{rush2015neural} and RNN~\cite{nallapati2016abstractive} as encoders and decoders. Furthermore, research endeavors focus on enhancing structures beyond the basic encoder-decoder framework. Hierarchical encoders~\cite{cheng2016neural, narayan2017neural, isonuma2017extractive, celikyilmaz2018deep, zhou2018neural} consist of an encoder for word-level encoding within sentences and another for encoding sentences into a document representation. Some studies propose layering CNN sentence encoders with LSTM document encoders and using LSTM decoders~\cite{cheng2016neural, narayan2017neural, isonuma2017extractive}. Additionally, research introduces FFNN as a decoder, incorporating a BiGRU-based sentence encoder and a document encoder~\cite{celikyilmaz2018deep, zhou2018neural}. Hsu et al.~\cite{hsu2018unified} integrate word- and sentence-level attention within the encoder-decoder model. Moreover, the bottom-up encoder-decoder model integrates word-level attention with sentence-level attention~\cite{gehrmann2018bottom}.

A mechanism-based text-to-text transformation approach has been introduced to improve the generation of natural and accurate summaries. Gu et al.~\cite{gu2016incorporating} proposed CopyNet, a sequence-to-sequence model with a copy mechanism to address the issue of UNKs~(Unknown words). The UNKs problem arises when words containing crucial information are omitted, limiting the vocabulary available for summarization to reduce computational costs. CopyNet was employed in a study to tackle the UNKs problem~\cite{gulcehre2016pointing}. Additionally, a method combining a pointer-generator and a coverage model was proposed to overcome the UNKs problem and to overcome repeating the same sentence problem~\cite{see2017get}. Another model, BART~(Bidirectional and Auto-Regressive Transformers)~\cite{lewis2020bart}, uses a sequence-to-sequence structure to mask part of the sentence, providing flexibility in learning. This improves noise flexibility and can be applied to arbitrary modifications of the original text. It is worth noting that BART is applied in both the mechanism-based and encoder-decoder frameworks.

Furthermore, instead of relying on an encoder-decoder framework, recent studies have explored the use of pre-training models that leverage large-scale data, such as Bidirectional Encoder Representations from Transformers~(BERT)~\cite{devlin2018bert} and Generative Pre-trained Transformer~(GPT)~\cite{radford2018improving}. BERT employs a bidirectional transformer encoder architecture, which captures contextual representations of all words in a sentence. This contextual information includes the relationships between each word and its surrounding words in the sentence. On the other hand, GPT utilizes a unidirectional transformer decoder to predict the next word based on the beginning part of a given sentence, focusing on understanding the contextual meaning and completing sentences using the sequential structure of the text. Learning methods such as BERT and GPT effectively capture text's semantic and structural characteristics by generating embeddings that represent sentences in a high-dimensional vector space. These embeddings have proven valuable for various text-to-text transformation tasks, including summarization, translation, and question-answering, as they implicitly encode the semantic information of the sentences.

\begin{table}[tbh]
    \setlength\tabcolsep{0pt}
    \setlength\thickmuskip{0mu}
    \setlength\medmuskip{0mu}
    \small
    \caption{\textcolor{black}{Datasets used for text-to-text transformation}}
    \centering
    \label{table:text2text_dataset}
    \begin{tabularx}{\textwidth}{p{0.1\textwidth}p{0.15\textwidth}p{0.45\textwidth}p{0.1\textwidth}p{0.1\textwidth}p{0.05\textwidth}p{0.05\textwidth}}
    \toprule
        \textbf{Category} & \textbf{Dataset} & \textbf{Description} & \centering \textbf{Open} & \textbf{Venue} & \textbf{Year} & \textbf{Ref} \\
    \midrule
        Keyword & Java-small & Java small dataset with 11 Java projects collected from GitHub & \centering O & ICPC & 2021 & \cite{ge2021keywords}\\
        \hline
        Keyword & self-collection & Chinese conversation crawling dataset from Baidu Tieba forum & \centering X & COLING & 2016 & \cite{mou-etal-2016-sequence} \\
        \hline
        Keyword & self-collection & Product review crawling dataset from jd.com, a Chinese e-commerce platform & \centering X & IEEM & 2017 & \cite{wang2017keyword} \\
        \hline
        Topic & TDT & 15,862 CNN news articles text dataset & \centering O & SIGIR & 1999 & \cite{hofmann1999probabilistic} \\
        \hline
        Topic & self-collection & Chinese dataset crawled 20 million comments from Baidu Tieba forum & \centering X & AAAI & 2017 & \cite{xing2017topic} \\
        \hline
        Topic & MXM & Official lyrics collection of the Million Song Dataset & \centering O & PMLR & 2017 & \cite{miao2017discovering} \\
        \hline
        Topic & 20NewsGroups & A dataset of approximately 18,000 newsgroup posts on 20 topics & \centering O & PMLR & 2017 & \cite{miao2017discovering} \\
        \hline
        Topic & self-collection & Corpora built to compare Chinese, English, and German on Wikipedia & \centering X & AAAI & 2019 & \cite{wei2019translating} \\
        \hline
        Generation & DUC & \begin{tabular}[c]{@{}l@{}}Document Understanding Conference~(DUC) dataset:\\manually and automatically summarized dataset of general\\documents\end{tabular} & \centering O & DUC & \begin{tabular}[c]{@{}l@{}}2000\\to \\2007\end{tabular} & \cite{ducdataset} \\
        \hline
        Generation & TAC & \begin{tabular}[c]{@{}l@{}}Text Analysis Conference~(TAC) dataset:\\knowledge base population, recognizing textual entailment, \\summarization, question answering\end{tabular} & \centering O & TAC & \begin{tabular}[c]{@{}l@{}}2008\\to\\2011\end{tabular} & \cite{tacdataset} \\
        \hline
        Generation & Gigaword & A dataset consisting of summaries created by combining the headlines and first sentences of articles & \centering O & EMNLP & 2015 & \cite{rush2015neural} \\
        \hline
        Generation & CNN/DM & A dataset consisting of summaries created using articles, questions, and answers from the Daily Mail & \centering O & CoNLL & 2016 & \cite{nallapati2016abstractive} \\
        \hline
        Generation & NYT & A dataset consisting of pairs of New York Times articles and human summaries & \centering O & NYPS & 2015 & \cite{hermann2015teaching} \\
        \hline
        Generation & WikiSum & A dataset containing Wikipedia articles, headlines, and cited sources & \centering O & AAAI & 2017 & \cite{nallapati2017summarunner} \\
        \hline
        Generation & Newroom & A dataset that contains news articles and metadata, such as headlines and human summaries & \centering O & ACL & 2016 & \cite{durrett2016learning} \\
        \hline
        Generation & C4 & An unlabeled dataset that refines text collected from large-scale crawls on the web & \centering O & JMLR & 2020 & \cite{raffel2020exploring} \\
    \bottomrule
    \end{tabularx}
\end{table}

\subsubsection{Dataset}
Text-to-text transformation models that rely on training require extensive datasets. Table~\ref{table:text2text_dataset} presents the datasets for the techniques introduced in this section. However, it is worth noting that the datasets used for extracting keywords and topics are not widely used due to their independent collection methods, as indicated by various research studies. In contrast, the dataset employed for the summarization technique is more universally utilized.

\subsubsection{Discussion and Challenges}
\label{dis:text2text}
Statistical methods for keyword and topic extraction suffer from constraints that necessitate customization based on document length or prior distribution of documents~\cite{cao2009density, armouty2019automated}. Additionally, their dependence on predefined rules presents a challenge when extracting semantic keywords or topics accurately. Text summarization has been approached through various methods, encompassing rule-based and template-based approaches, encoder-decoders, transformers, BERTs, and GPTs. However, despite these diverse techniques, the challenge of producing summaries with sufficient variety and naturalness persists.

One approach to address these challenges in text-to-text transformation involves incorporating external knowledge sources, such as lexical chains and knowledge graphs. Alternatively, pre-training models trained on large-scale data, such as BERTs and GPTs, can be utilized~\cite{yu2022survey}. However, using learning models to achieve the highest performance remains challenging, primarily due to the difficulty in interpreting how the text was generated, posing interpretability issues. To tackle these obstacles, researchers may need to explore innovative ways to combine different techniques effectively and enhance the interpretability of the models.
\subsection{Image-to-Text Transformation}
\label{subsec:image2text}

\begin{figure*}[b]
	\centering
	\includegraphics[width=.8\textwidth]{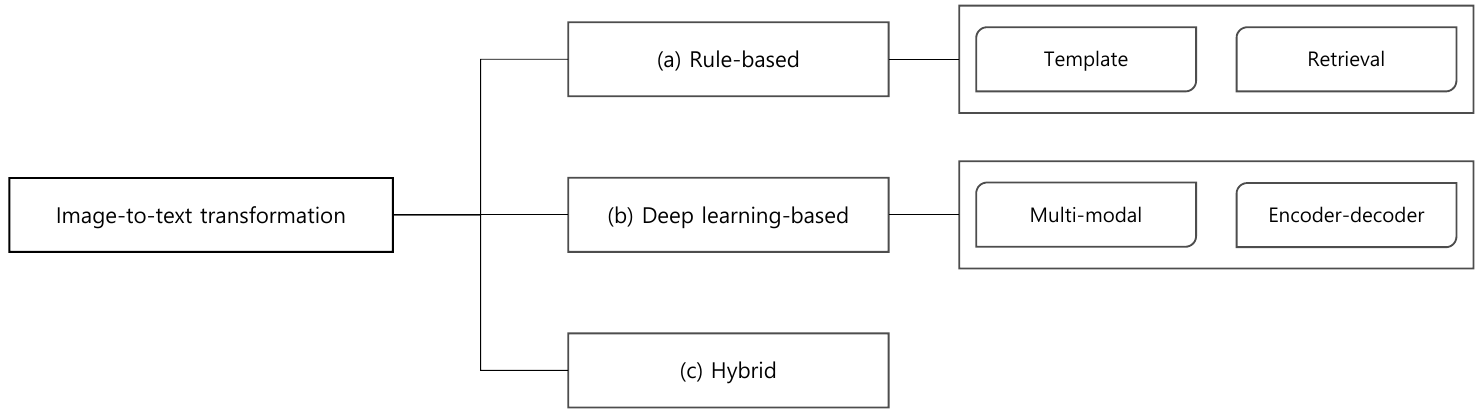}
	\caption{Strategies of image-to-text. (a) is rule-based, (b) is deep learning, and (c) is a hybrid strategy.}
	\label{F:img_to_text}
\end{figure*}

Image-to-text transformation involves generating textual descriptions for entire images or specific parts of an image. It has a similar process to automatic image indexing, which is the process of automatically indexing images to make it easier to classify or search~\cite{hossain2019comprehensive}. For example, in the feature extraction process, methods such as Convolutional Neural Networks (CNN) are used to extract image characteristics and apply pooling techniques. Image-to-text transformations are finding applications in diverse domains like e-commerce~\cite{s23031286,9897417} and tourism~\cite{Fudholi_2021,9286602}. Before image-to-text conversion, feature extraction is performed to utilize the essential characteristics of the image data. The overall strategy for converting image data and its extracted features into text is illustrated in Figure~\ref{F:img_to_text}, 
presenting three main approaches for image-to-text transformation (a) rule-based, (b) deep learning, and (c) hybrid methods.

\subsubsection{Rule-based Method}
Rule-based methods in image-to-text transformation generate straightforward text, such as descriptive captions, to help understand the image content. These methods are categorized into retrieval-based and template-based models. In retrieval-based methods, texts are generated by selecting a sentence from a predefined pool of sentences or by combining multiple sentences from the pool. The sentence pool technique utilizes a model to generate highly relevant text to the input image~\cite{mason2014nonparametric, yagcioglu2015distributed, verma2014im2text}. For instance, Mason et al.~\cite{mason2014nonparametric} employ a word frequency model to find smoothed-estimated captions for the contents of the input image among a collection of image caption data. The SumBasic algorithm~\cite{nenkova2005impact} is used to maximize the occurrence of high-frequency words among the words included in candidate sentences. Yagcioglu et al.~\cite{yagcioglu2015distributed} utilize a neural network-based language model to find visually similar images to the input image and then use a distributional model of meaning for sentences. The model for finding similar images obtains the average caption of visually similar images using a visual query. The frequency model generates text by averaging the captions of images similar to the input image. Verma et al.~\cite{verma2014im2text} create text describing image by learning a bi-directional association between image data and text data based on an SVM-based structural model. Since structural SVM performs both image retrieval and recognition, it functions similarly to the mechanism of rule-based retrieval.

The template-based method converts detected objects within input images into text by fitting them into sentence templates. Consequently, this method requires techniques for object detection and matching objects to sentence templates~\cite{yang2011corpus, li2011composing, mitchell2012midge}. Yang et al.~\cite{yang2011corpus} employ the PASCAL-VOC 2008 dataset for object detection in images. The detected objects are mapped to sentence templates using classes set by the authors, encompassing objects, actions, scenes, prepositions, and a hidden Markov Model, thereby transforming them into text. Li et al.~\cite{li2011composing} utilize RBF kernel SVM for object detection, extracting features from images such as objects, visual attributes, and spatial relationships. They transform these features, collected from web-scale n-grams, into text using a phrase fusion composition algorithm. Mitchell et al.~\cite{mitchell2012midge} employ a method filtering co-occurrence statistics attributes. The text transformation adheres to Treebank parsing guidelines.

\subsubsection{Deep Learning-based Method}
The deep learning-based approach learns from extensive datasets to convert a broader range of images into text. Among the models used for image-to-text transformation in deep learning, there are multimodal learning models and encoder-decoder framework models. In these methods, CNNs are primarily used to extract visual features at various levels within images, while RNNs are often employed to process sequential text data~\cite{baltrusaitis2019multimodal}. 

In multimodal learning, the model is trained by embedding image and text data into different layers~\cite{mao2014explain, karpathy2015deep, chen2015mind}. The idea behind multimodal learning is effectively combining methods for processing these different data types. Given the challenge of interpreting new images depicting previously unseen scenes, multimodal learning aids in enhancing performance by modeling the probability distribution for word generation, taking into account both the images and preceding words. For instance, Mao et al.~\cite{mao2014explain} proposed the m-RNN model, which consists of text processing, image processing, and a multimodal layer based on multimodal Recurrent Neural Networks~(m-RNN). The text processing layer learns feature embeddings for words in a dictionary and captures semantic and temporal context using a recurrent layer. The image processing layer involves CNN layers to extract image features, and the multimodal layer connects language models and CNNs. Similarly, Karpathy et al.~\cite{karpathy2015deep} proposed a model with a structure similar to m-RNN, where the text processing layer is Bi-RNN, and the image processing layer comprises CNN. Chen et al.~\cite{chen2015mind} proposed a model based on Bi-RNN that enables bidirectional representations between images and image descriptions, allowing visual representations to incorporate new information in creating and comprehending text.

On the other hand, the encoder-decoder framework consists of an encoder module for image compression and a decoder module for text generation~\cite{vinyals2015show, donahue2015long, you2016image}. The encoder creates a context vector that compresses the image information, and the decoder sequentially generates words related to the context vector. For example, Vinyals et al.~\cite{vinyals2015show} proposed a model with a CNN encoder and RNN decoder. In contrast, Donahue et al.~\cite{donahue2015long} proposed a model with a CNN encoder and LSTM decoder structure. You et al.~\cite{you2016image} introduced a model that uses CNN, RNN, and attention mechanisms to improve accuracy and speed. CNN extracts visual features from images. RNNs and attention mechanisms capture salient content and then convert it into text. This process enhances accuracy and speed by focusing on converting only important salient content into text instead of processing the entire image information.

\subsubsection{Hybrid Method}
Hybrid methods combine the advantages of both rule-based and deep learning approaches. These strategies leverage deep learning to extract features or concepts from images and then utilize rule-based methods to structure these features into a coherent text. For instance, Ma et al.~\cite{ma2015multimodal} proposed a multimodal convolutional neural network framework that utilizes a neural network for image-text matching and score calculation, following the retrieval methods. Lebret et al.~\cite{lebret2014simple} use a CNN and a purely bilinear model to predict syntactic components likely to describe a given image and then convert them into text using a straightforward statistical language model. Yan et al.~\cite{yan2015deep} match images and texts in a joint latent space learned through deep canonical correlation analysis, addressing complexity and overfitting challenges of previously proposed methods. These hybrid methods efficiently combine deep learning capability to extract relevant features with rule-based techniques to generate accurate and meaningful textual descriptions from images.

\subsubsection{Dataset}
Table~\ref{table:img2text_dataset} shows the datasets utilized in the techniques described in this section for image-to-text transformation. The datasets used for image-to-text technology encompass PASCAL-VOC, SBU Captions, Flickr8K, Flickr30k, and MS-COCO. These datasets contain image-text pairs, which the model uses to learn many transformations.
\begin{table}[!htbp]
    \setlength\tabcolsep{0pt}
    \setlength\thickmuskip{0mu}
    \setlength\medmuskip{0mu}
    \small
    \caption{\textcolor{black}{Datasets used for image-to-text transformation}}
    \centering
    \label{table:img2text_dataset}
    \begin{tabularx}{\textwidth}{p{0.18\textwidth}p{0.56\textwidth}p{0.08\textwidth}p{0.08\textwidth}p{0.05\textwidth}p{0.05\textwidth}}
    \toprule
        \textbf{Dataset} & \textbf{Description} & \centering \textbf{Open} & \textbf{Venue} & \textbf{Year} & \textbf{Ref} \\
    \midrule
        PASCAL-VOC & A dataset labeled with captions for objects, actions, and places in images and images collected in the PASCAL VOC Challenge & \centering O & IJCV & 2010 & \cite{everingham2010pascal} \\
        \hline
        SBU Captions & A dataset labeled with user images and user descriptions collected from Flickr & \centering O & NIPS & 2011 & \cite{ordonez2011im2text} \\
        \hline
        Flickr8K & A dataset of 8,000 images of people participating in sports activities, labeled with five descriptions of the images & \centering O & JAIR & 2013 & \cite{hodosh2013framing} \\
        \hline
        Flickr30k & A dataset labeled with 30K images and five descriptions of the photos for everyday activities, events, and scenes & \centering O & TACL & 2014 & \cite{young2014image} \\
        \hline
        MS-COCO & A dataset consisting of images and descriptions of images containing 91 objects that are easy for a 4-year-old child to distinguish & \centering O & ECCV & 2014 & \cite{lin2014microsoft} \\
        \hline
        UIUC Pascal Sentence & A crowdsourced, quality-controlled dataset of 9,000 images and 40,000 descriptions & \centering O & NAACL-HLT & 2010 & \cite{rashtchian2010collecting} \\
    \bottomrule
    \end{tabularx}
\end{table}

\subsubsection{Discussion and Challenges}
\label{dis:img2text}
Previous studies in image-to-text transformation have primarily focused on object recognition and expressing object relationships. However, achieving natural and accurate image-to-text transformation requires additional external knowledge. To address these limitations, researchers have proposed methods such as creating personalized text by incorporating the user's prior knowledge, active vocabulary, and writing style~\cite{zhang2020learning}, and controllable captioning methods that allow users to directly intervene in the text generation process~\cite{cornia2019show, deng2020length}. Moreover, the current research in image-to-text transformation has limitations related to representative datasets. The publicly available datasets and proposed models may not be sufficient for the broader application of image-to-text technology in various medical, manufacturing, and academic domains. Issues arise when object detection models fail to identify specific objects, resulting in information loss during image-to-text transformation. Addressing these challenges involves: handling ambiguous or complex images, generating text in diverse styles or tones, and improving the interpretability of the models.
\subsection{Video-to-Text Transformation}
\label{subsec:video2text}

\begin{figure*}
    \centering
    \includegraphics[width=.9\textwidth]{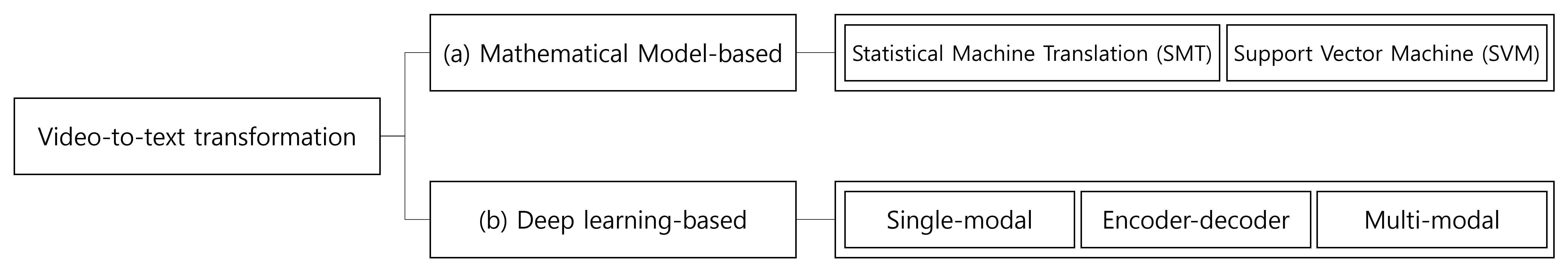}
    \caption{Strategies of video-to-text transformation. Video-to-text is classified into (a) mathematical model-based method and (b) deep learning-based method.}
    \label{F:video_to_text}
\end{figure*}

Similar to the image-to-text transformation discussed in Section~\ref{subsec:image2text}, the process of video-to-text transformation aims to translate videos into coherent sentences that capture the context and dynamic elements, including object movements. The video-to-text process involves extracting relevant information from the video and generating descriptive text, resulting in a caption describing the video content. Figure~\ref{F:video_to_text} illustrates strategies of video-to-text transformation. Figure~\ref{F:video_to_text}~(a) is a mathematical model-based method, and (b) is a deep learning-based method.

\subsubsection{Mathematical Model-based Methods}
This method involves extracting visual information from the video and generating sentence-like descriptions based on the extracted data. These models employ operations like prediction, classification, and clustering to understand elements and patterns within the video input. However, one of the challenges of using mathematical models for natural language conversion is learning an extensive vocabulary. Various mathematical models, such as Support Vector Machine~(SVM) and Statistical Machine Translation~(SMT), adopt sentence structure detection techniques to process the video data~\cite{rohrbach2013translating, guadarrama2013youtube2text}. For example, Rohrbach et al.~\cite{rohrbach2013translating} employ SMT to extract SVO video information from the TACos dataset~\cite{regneri2013grounding}, generating natural language descriptions that include activities, tools, objects, sources, and targets. They classify leaf nodes containing object details using a non-linear SVM based on the extracted SVO built in a tree format. Subsequently, the zero-shot language model assesses and ranks the generated candidate sentences based on their plausibility. Guadarrama et al.~\cite{guadarrama2013youtube2text} derive a semantic representation~(SR) from video content to articulate visual perception in natural language. They obtain SR from the video data using a Conditional Random Field~(CRF). Subsequently, they employ SMT to describe the SR. Despite their capabilities, these mathematical models have limitations in achieving optimal performance for video processing due to the restricted amount of data they can effectively learn.

\subsubsection{Deep Learning-based Method}
As depicted in Figure~\ref{F:video_to_text}~(b), the deep learning-based approach for video-to-text transformation comprises three main components: single-modal, encoder-decoder, and multi-modal. These methods address the limitations of mathematical models that struggle with learning extensive vocabulary.

In the single-modal method, a model involves generating text directly from specific features extracted directly from the video, aiming to establish a clear connection between video elements and linguistic representation~\cite{xu2016msr, tanaka2021lol}. For instance, Xu et al.~\cite{xu2016msr} generate video captions based on motion information extracted from the MSR-VTT~\cite{xu2016msr} dataset using 3D~CNN. Similarly, Venugopalan et al.~\cite{venugopalan2015sequence} demonstrate this by generating video captions through an LSTM model that learns the temporal structure from the sequence of frames in the M-VAD~\cite{torabi2015using} and MPII-MD~\cite{rohrbach2015dataset} datasets.

In the encoder-decoder framework, the encoder extracts visual features from the video, while the decoder converts these visual features into descriptive text. Yang et al.~\cite{yang2023vid2seq} proposed a model consisting of a transformer-based visual encoder, text encoder, and text decoder to generate contextually rich video captions by integrating visual and audio information. Tanaka and Simo-Serra~\cite{tanaka2021lol} extract gameplay elements from the LOL-V2T~\cite{tanaka2021lol} dataset using CNN and then generate game video captions using various models such as a vanilla transformer~\cite{zhou2018end}, MART-based captioning model~\cite{lei2020mart}, and RNN decoder.

In the multi-modal approach, a multi-modal transformer is utilized for dense video captioning. Zhu et al.~\cite{zhu2022end} employed a multi-modal transformer and a decoder. The transformer assimilates information from the preceding caption and the present frame to craft the following caption. Captions are then generated by concurrently determining event locations and their associated descriptions using the decoder. In summary, deep learning-based methods for video-to-text transformation encompass directly utilizing visual features, transformation through context extraction, and various other innovative approaches.

\subsubsection{Dataset}
Table~\ref{T:video2text_dataset} presents the datasets utilized for video-to-text transformation techniques. These datasets consist of pairs of videos and corresponding textual descriptions that provide a comprehensive understanding of the video content. Furthermore, each dataset is associated with a specific context, such as cooking, movie, or video game, and in some cases, it may include multiple contexts.

\begin{table}[!htbp]
    \small
    \caption{\textcolor{black}{Dataset used for video-to-text transformation}}
    \centering
    \label{T:video2text_dataset}
    \begin{tabular}{p{0.2\textwidth}p{0.05\textwidth}p{0.12\textwidth}p{0.05\textwidth}p{0.07\textwidth}p{0.08\textwidth}p{0.08\textwidth}p{0.08\textwidth}p{0.05\textwidth}}
    \toprule
        \multicolumn{1}{c}{\textbf{Dataset}}  & \multicolumn{1}{c}{\textbf{Year}} & \multicolumn{1}{c}{\textbf{Context}} & \multicolumn{1}{c}{\textbf{\#Video}} & \multicolumn{1}{c}{\textbf{\#Clip}} & \multicolumn{1}{c}{\textbf{\#Sentence}} & \multicolumn{1}{c}{\textbf{\#Word}} & \multicolumn{1}{c}{\textbf{Vocabulary}} & \multicolumn{1}{c}{\textbf{Ref}}\\
    \midrule
        MSVD & \multicolumn{1}{c}{2011} & multi-category & \multicolumn{1}{c}{-} & \multicolumn{1}{r}{1,970} & \multicolumn{1}{r}{70,028} & \multicolumn{1}{r}{607,339} & \multicolumn{1}{r}{13,101} & \multicolumn{1}{c}{\cite{chen2011collecting}}   \\
        TACoS & \multicolumn{1}{c}{2013} & cooking & \multicolumn{1}{r}{127} & \multicolumn{1}{r}{7,206} & \multicolumn{1}{r}{18,227} & \multicolumn{1}{c}{-} & \multicolumn{1}{c}{-} & \multicolumn{1}{c}{\cite{regneri2013grounding}} \\
        TACos Multi-Level & \multicolumn{1}{c}{2014} & cooking & \multicolumn{1}{r}{185} & \multicolumn{1}{r}{14,105} & \multicolumn{1}{r}{52,593} & \multicolumn{1}{c}{-} & \multicolumn{1}{c}{-} & \multicolumn{1}{c}{\cite{rohrbach2014coherent}} \\
        M-VAD & \multicolumn{1}{c}{2015} & movie & \multicolumn{1}{r}{92} & \multicolumn{1}{r}{48,986} & \multicolumn{1}{r}{55,904} & \multicolumn{1}{r}{519,933} & \multicolumn{1}{r}{18,269} & \multicolumn{1}{c}{\cite{torabi2015using}}      \\
        MPII-MD & \multicolumn{1}{c}{2015} & movie & \multicolumn{1}{r}{94} & \multicolumn{1}{r}{68,337} & \multicolumn{1}{r}{68,375} & \multicolumn{1}{r}{653,467} & \multicolumn{1}{r}{24,549} & \multicolumn{1}{c}{\cite{rohrbach2015dataset}}  \\
        MSR-VTT & \multicolumn{1}{c}{2016} & 20 categories & \multicolumn{1}{r}{7,180} & \multicolumn{1}{r}{10,000} & \multicolumn{1}{r}{200,000} & \multicolumn{1}{r}{1,856,523} & \multicolumn{1}{r}{29,316} & \multicolumn{1}{c}{\cite{xu2016msr}} \\
        Charades & \multicolumn{1}{c}{2016} & human & \multicolumn{1}{r}{30} & \multicolumn{1}{r}{9,848} & \multicolumn{1}{r}{27,847} & \multicolumn{1}{c}{-} & \multicolumn{1}{c}{-} & \multicolumn{1}{c}{\cite{sigurdsson2016hollywood}} \\
        ActivityNet captions & \multicolumn{1}{c}{2017} & open & \multicolumn{1}{r}{20,000} & \multicolumn{1}{r}{100,000} & \multicolumn{1}{r}{100,000} & \multicolumn{1}{c}{-} &  \multicolumn{1}{c}{-} & \multicolumn{1}{c}{\cite{krishna2017dense}}  \\
        YooCook2 & \multicolumn{1}{c}{2018} & cooking & \multicolumn{1}{r}{2,000} & \multicolumn{1}{r}{14,000} & \multicolumn{1}{r}{14,000} & \multicolumn{1}{c}{-} & \multicolumn{1}{r}{2,600} & \multicolumn{1}{c}{\cite{zhou2018towards}}  \\
        EPIC-KITCHENS & \multicolumn{1}{c}{2018} & cooking & \multicolumn{1}{c}{-} & \multicolumn{1}{r}{115,000} & \multicolumn{1}{r}{39,596} & \multicolumn{1}{c}{-} & \multicolumn{1}{c}{-} & \multicolumn{1}{c}{\cite{damen2018scaling}}  \\
        HowTo100M & \multicolumn{1}{c}{2019} & instruction & \multicolumn{1}{r}{1,221} & \multicolumn{1}{r}{1,360,000} & \multicolumn{1}{r}{1,360,000} & \multicolumn{1}{c}{-} & \multicolumn{1}{c}{-} & \multicolumn{1}{c}{\cite{miech2019howto100m}}  \\
        Getting Over It & \multicolumn{1}{c}{2019} & video game & \multicolumn{1}{r}{8} & \multicolumn{1}{r}{9,700} & \multicolumn{1}{r}{63,000} & \multicolumn{1}{c}{-} & \multicolumn{1}{r}{2,598} & \multicolumn{1}{c}{\cite{li2019end}}  \\
        LOL-V2T & \multicolumn{1}{c}{2021} & video game & \multicolumn{1}{r}{157} & \multicolumn{1}{r}{9,723} & \multicolumn{1}{r}{62,677} & \multicolumn{1}{c}{-} & \multicolumn{1}{c}{-} & \multicolumn{1}{c}{\cite{tanaka2021lol}} \\
    \bottomrule
    \end{tabular}
\end{table}

\subsubsection{Discussion and Challenges}
\label{dis:video2text}
Video-to-text transformation generates text that involves a sequence of images from video frames. However, most existing video-to-text studies focus on short-term videos, making it challenging to maintain context across multiple frames over longer durations~\cite{guadarrama2013youtube2text, krishna2017dense}. There are problems with tracking an object's action over a long time or multiple objects. As a result, one of the primary challenges is providing comprehensive descriptions encompassing multiple scenes or actions within the video. Expressing actions or multiple scenes spanning frames in a single sentence is particularly challenging~\cite{rohrbach2013translating}, necessitating research on complex sentence generation models. In order to enable long-term video-to-text transformation, datasets that include longer video clips are essential. Another issue arises from the lack of clarity in defining how to aggregate video frames into a single sentence. Researchers are actively working to address this problem and develop appropriate methods for video summarization. Moreover, a video-text dataset that includes a sufficient number of diverse activities and sentences is essential to enhance the performance of the models.
\section{Data-to-Graph Transformation}
\label{sec:d2g}
The graph format data offers the advantage of providing a structured representation and capturing abstract relationships between data for learning purposes. Data structuring involves composing text in natural language or modeling object relations in images and videos to understand the underlying connections between data. Abstract relationships between data are primarily inferred through graph neural network learning models. These models utilize the structured data, represented as graphs, as input to learn object relations or semantic information. This section explores transformation strategies for data from text, image, and video formats into graph format, leveraging the benefits of graph representation.

\subsection{Text-to-Graph Transformation}
\label{subsec:text2graph}

\begin{figure*}
    \centering
    \includegraphics[width=.9\textwidth]{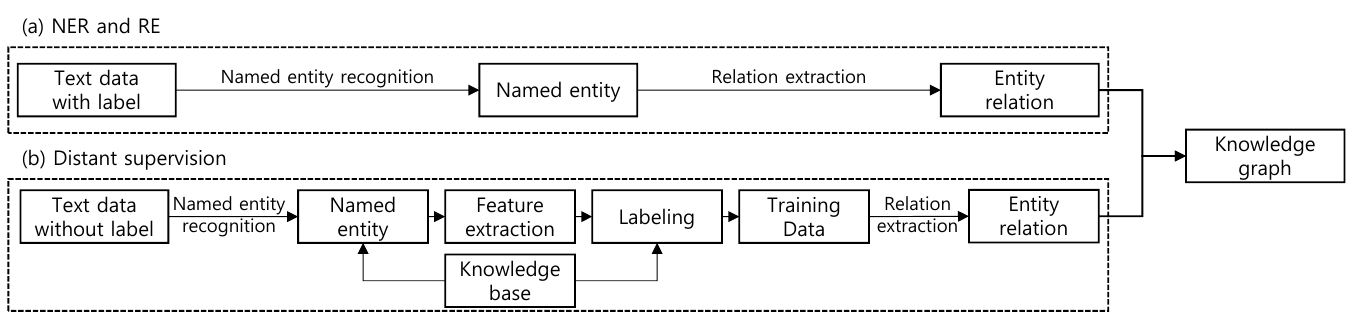}
    \caption{Strategies of text-to-graph transformation. Text-to-graph transformation strategies include (a) Named Entity Recognition~(NER) and Relation Extraction~(RE) and (b) distant supervision techniques.}
    \label{F:text_to_graph}
\end{figure*}

Text-to-graph involves converting text data into a Knowledge Graph (KG). Text data encompasses various types of information. For instance, as introduced in Section~\ref{subsec:table2text}, text generated from table-to-text techniques inherently contains relationships between table and entity data. A KG represents relationships among real-world entities such as objects, events, and concepts in a graph structure. KG construction involves identifying embedded entities within text and extracting relations, known as information extraction. Figure~\ref{F:text_to_graph} illustrates strategies for transforming text-format data into a KG. Figure~\ref{F:text_to_graph}~(a) demonstrates the information extraction process using Named Entity Recognition (NER) and Relation Extraction (RE). (b) shows the process of resolving labeling inaccuracies in the data using distant supervision and extracting relations.

\subsubsection{NER and RE Method}
Named entities are specific data elements with special meaning, such as people, places, organizations, and scores, inherent in the text, as illustrated in Figure~\ref{F:text_to_graph}~(a). NER identifies these entities in the text, while RE discovers the relationships between named entities identified through NER. Text-to-graph techniques utilizing NER and RE employ different effective methods depending on the domain, leading to the proposal of various approaches. NER and RE information can be extracted through rule-based traditional methods or learning-based models. The traditional NER and RE methods extract named entities and relations based on user-defined rules~\cite{liu_2019_t2g, xu_2020_t2g}. For example, in the stock price prediction domain, Liu et al.~\cite{liu_2019_t2g} focused on sentiments in news that influence stock prices. They transformed unstructured text into a KG by creating tuples like \textit{(Company A, shareholding, Company B)}, which served as input data for the Gated Recurrent Unit~(GRU) model. Additionally, a tree kernel-based RE was proposed~\cite{zelenko2003kernel, zhou2005exploring, sun_2014_t2g}. The tree kernel-based RE extracts relations between entities using similarity measures of a syntactic parse tree generated from an unstructured natural language. Han and Sun~\cite{sun_2014_t2g} introduced a feature-enriched tree kernel to capture semantic relation information between two entities, going beyond the traditional syntactic tree representation~\cite{zelenko2003kernel, zhou2005exploring}.

The learning-based model offers a cost-effective approach for generating KG compared to traditional NER and RE methods, as it reduces the dependency on human-annotated resources. Kim et al.~\cite{kim_2021_t2g} introduced CO-BERT, a BERT-based model for extracting REs focusing on the COVID-19 dataset, and an HRT structure for generating the KG. They employed entity dictionaries to filter COVID-19 wiki data and sentences obtained from COVID-19 news data. These dictionaries and sentences were then used as inputs for the CO-BERT model to generate CO-BERT features, which were further converted into KG using HBT-based RE. Xu et al.~\cite{xu_2020_t2g} concentrated on establishing NER and RE connections between bio-entities, authors, articles, affiliations, and funding. They proposed BioBERT, a BERT-based model, to generate a KG that facilitates searching for papers in PubMed and PMC datasets. In addition, Chang et al.~\cite{chang_2014_t2g} presented a tensor decomposition approach to enhance RE performance based on a learning model. Rotmensch et al.~\cite{rotmensch_2017_t2g} investigated the direct linkage between diseases and symptoms in de-identified patient emergency department records. They employed the maximum likelihood of logistic regression, naive Bayesian classifier, and Bayesian network with noisy OR gates to generate knowledge graphs automatically. Chen et al.~\cite{chen_2018_t2g} extracted data from textbooks, curriculum standards, and course tutorials and then used Optical Character Recognition~(OCR) to create a KG. The KG is constructed by identifying relationships between concepts through probabilistic association rule mining.

\subsubsection{Distant Supervision Method}
Figure~\ref{F:text_to_graph}~(b) illustrates distant supervision as a labeling method that uses a knowledge base database to estimate the relationship between two entities in the RE process without specific target labels. One of the significant challenges in the RE process is the potential omission of multiple relations, as relationships are typically extracted only within a single sentence.

To address this limitation, Jiang et al.~\cite{jiang_2016_t2g} introduced a learning-based model that includes sentence-level feature extraction, cross-sentence max-pooling, and multi-label relation modeling to capture multiple relations. Similarly, Lin et al.~\cite{lin_2016_t2g} and Zeng et al.~\cite{zeng_2017_t2g} employed a sentence-level attention-based model to prevent mislabeling during the distant supervision-based RE process. Angeli et al.~\cite{angeli_2014_t2g} improved performance by applying partial supervision to distant supervision-based RE. Han and Sun~\cite{han_2014_t2g} utilized semantic consistency to enable multi-instance and multi-label learning. Pershina et al.~\cite{pershina_2014_t2g} proposed a multi-instance multi-label model that performs RE using human-labeled data. Chen et al.~\cite{chen_2014_t2g_clue} addressed the problem of disregarding global clues from the knowledge base in distant supervision. They introduced a collaborative inference framework to reconcile discrepancies between local predictions using global clues.

\subsubsection{Dataset}
Text-to-graph transformation relies on domain-specific datasets to extract focused information for a given domain. Consequently, different datasets are utilized based on the domain and objectives of each study. Table~\ref{table:t2g_dataset} summarizes the datasets used in the text-to-graph transformation techniques discussed in Section~\ref{subsec:text2graph}.

\begin{table}[!htbp]
    \small
    \caption{Dataset for text-to-graph transformation}
    \centering
    \label{table:t2g_dataset}
    \begin{tabular}    {p{0.2\textwidth}p{0.1\textwidth}p{0.1\textwidth}p{0.1\textwidth}p{0.13\textwidth}p{0.1\textwidth}p{0.1\textwidth}p{0.12\textwidth}p{0.05\textwidth}}
        \toprule
            \multicolumn{1}{c}{\textbf{Dataset}} & \multicolumn{1}{c}{\textbf{\#Sentence}} & \multicolumn{1}{c}{\textbf{\#Word}} & \multicolumn{1}{c}{\textbf{\#Token}} & \multicolumn{1}{c}{\textbf{\#Entity pair}} & \multicolumn{1}{c}{\textbf{\#Fact}} & \multicolumn{1}{c}{\textbf{\#Triple}} & \multicolumn{1}{c}{\textbf{\#Document}} & \multicolumn{1}{c}{\textbf{Ref}} \\
        \midrule
            CORD-19 & \multicolumn{1}{c}{-} & \multicolumn{1}{c}{-} & \multicolumn{1}{c}{-} & \multicolumn{1}{c}{-} & \multicolumn{1}{c}{-} & \multicolumn{1}{c}{-} & \multicolumn{1}{r}{1,000,000} & \multicolumn{1}{c}{-} \\
            PubMed & \multicolumn{1}{c}{-} & \multicolumn{1}{c}{-} & \multicolumn{1}{c}{-} & \multicolumn{1}{c}{-} & \multicolumn{1}{c}{-} & \multicolumn{1}{c}{-} & \multicolumn{1}{r}{44,338} & \multicolumn{1}{c}{-} \\
            NELL & \multicolumn{1}{c}{-} & \multicolumn{1}{c}{-} & \multicolumn{1}{c}{-} & \multicolumn{1}{r}{753,000} & \multicolumn{1}{c}{-} & \multicolumn{1}{r}{1,800,000} & \multicolumn{1}{c}{-} & \multicolumn{1}{c}{\cite{NELL}} \\
            ACE RDC 204 & \multicolumn{1}{c}{-} & \multicolumn{1}{c}{-} & \multicolumn{1}{c}{-} & \multicolumn{1}{c}{-} & \multicolumn{1}{r}{5,702} & \multicolumn{1}{c}{-} & \multicolumn{1}{r}{451} & \multicolumn{1}{c}{\cite{ACERDC}} \\
            NYT 10 & \multicolumn{1}{r}{695,059} & \multicolumn{1}{c}{-} & \multicolumn{1}{c}{-} & \multicolumn{1}{r}{377,948} & \multicolumn{1}{r}{20,202} & \multicolumn{1}{c}{-} & \multicolumn{1}{c}{-} & \multicolumn{1}{c}{\cite{NYT10}} \\
            \begin{tabular}[c]{@{}l@{}}NYT based\\ new dataset\end{tabular} & \multicolumn{1}{r}{1,117,786} & \multicolumn{1}{c}{-} & \multicolumn{1}{c}{-} & \multicolumn{1}{r}{509,115} & \multicolumn{1}{r}{61,396} & \multicolumn{1}{c}{-} & \multicolumn{1}{c}{-} & \multicolumn{1}{c}{-} \\
            KBP & \multicolumn{1}{c}{-} & \multicolumn{1}{c}{-} & \multicolumn{1}{c}{-} & \multicolumn{1}{r}{3,334} & \multicolumn{1}{r}{186,396} & \multicolumn{1}{c}{-} & \multicolumn{1}{r}{1,500,000} & \multicolumn{1}{c}{\cite{KBP}} \\
            DBpedia & \multicolumn{1}{r}{187,000} & \multicolumn{1}{c}{-} & \multicolumn{1}{c}{-} & \multicolumn{1}{r}{80,000} & \multicolumn{1}{c}{-} & \multicolumn{1}{c}{-} & \multicolumn{1}{c}{-} & \multicolumn{1}{c}{\cite{DBpedia}} \\
            HudongBaike & \multicolumn{1}{r}{203,000} & \multicolumn{1}{c}{-} & \multicolumn{1}{c}{-} & \multicolumn{1}{r}{100,000} & \multicolumn{1}{c}{-} & \multicolumn{1}{c}{-} & \multicolumn{1}{c}{-} & \multicolumn{1}{c}{-} \\
            Googlee Billion Word & \multicolumn{1}{r}{30,301,028} & \multicolumn{1}{r}{800,000} & \multicolumn{1}{r}{1,000M} & \multicolumn{1}{c}{-} & \multicolumn{1}{c}{-} & \multicolumn{1}{c}{-} & \multicolumn{1}{c}{-} & \multicolumn{1}{c}{\cite{GoogleBillionWord}}  \\
            WikiText-103 & \multicolumn{1}{r}{28,595} & \multicolumn{1}{r}{200,000} & \multicolumn{1}{r}{100M} & \multicolumn{1}{c}{-} & \multicolumn{1}{c}{-} & \multicolumn{1}{c}{-} & \multicolumn{1}{c}{-} & \multicolumn{1}{c}{\cite{WikiText103}} \\
            CNCSM & \multicolumn{1}{r}{1,847} & \multicolumn{1}{r}{36,697} & \multicolumn{1}{c}{-} & \multicolumn{1}{c}{-} & \multicolumn{1}{c}{-} & \multicolumn{1}{c}{-} & \multicolumn{1}{c}{-} & \multicolumn{1}{c}{-} \\
        \bottomrule
    \end{tabular}
\end{table}

\subsubsection{Discussion and Challenges}
\label{dis:text2graph}
The text-to-graph transformation technique represents the connections between real-world entities in a graph, rendering it suitable for training artificial intelligence models tailored to domain-specific problem-solving. Nevertheless, text-to-graph transformation does possess certain limitations that hinder its universal applicability. The information extraction required from the data varies depending on the domain. As seen in Table~\ref{table:t2g_dataset}, the datasets used for specific purposes are distinct, highlighting the challenge of low scalability in this context. Moreover, challenges arise from the diverse languages of the collected data, the absence of a robust pre-trained language model for extracting relationships from text, and the task of identifying relationships amidst inconsistent word order. These challenges hinder the efficient use of current datasets. Consequently, research should focus on developing language-agnostic RE systems to advance text-to-graph techniques, leveraging transfer learning to use NER models in different languages, and creating high-efficacy dictionary learning models to extract inter-word relationships.
\subsection{Image-to-Graph Transformation}
\label{subsec:img2graph}

\begin{figure*} [b]
    \centering
    \includegraphics[width=.9\textwidth]{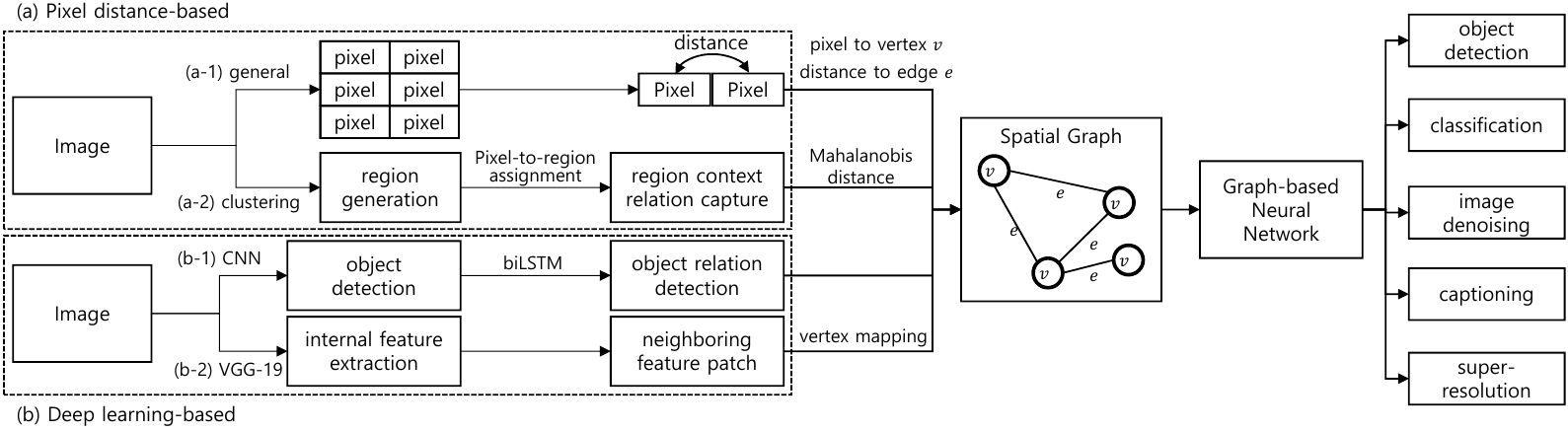}
    \caption{Strategies of image-to-graph. (a) is pixel distance-based method and (b) is deep learning-based method.
    }
    \label{F:image_to_graph}
\end{figure*}

Image-to-graph transformation entails identifying objects within an image and establishing relationships between these objects to create a graph representation. This transformed graph can be utilized in graph-based deep learning models to depict object relationships, offering an advantage in capturing image topology compared to prevalent computer vision-based deep learning models~\cite{prabhu_2015_i2g}. Consequently, image-to-graph transformation has attracted considerable attention across various domains, including classification, denoising, super-resolution, and captioning. Figure~\ref{F:image_to_graph} illustrates diverse strategies for image-to-graph transformation, featuring (a) and (b) representing pixel distance methods and deep learning-based approaches, respectively.

\subsubsection{Pixel Distance-based Method}
The pixel distance-based method transforms an image into a graph by considering each pixel in the image as a vertex and establishing edges based on pixel distances. In this context, the vertices and edges are crucial in pixel distance-based techniques. Figure~\ref{F:image_to_graph}~(a-1) illustrates a typical pixel distance-based method. For instance, Hong et al.~\cite{hong_2021_i2g} investigated effectively identifying objects of interest in hyperspectral images by defining pixels as vertices and capturing similarities between pixels as edges. Prabhu and Babu~\cite{prabhu_2015_i2g} used a CNN to detect objects within an image and described each object as a vertex. They represented edge features as the distance between object pixels, the spatial organization between two objects, and the degree of overlap of the boundary boxes of the two objects. Valesia et al.~\cite{valesia_2020_i2g} also adopted the pixel distance-based approach. They defined each image pixel as a vertex and calculated the Euclidean distance between the feature vector of each pixel and the pixel feature vectors inside the search window. Based on these calculated distances, they created a k-regular graph connecting each pixel to its k-nearest neighbors. The resulting k-regular graph was then utilized as input for a graph-convolutional denoiser network, contributing to improving image denoising tasks.

Figure~\ref{F:image_to_graph}~(a-2) illustrates the pixel distance-based method based on clustering. Wan et al.~\cite{wan_2021_i2g} employ a simple linear iterative clustering algorithm to divide the image into regions effectively. The pixel-to-region assignment is learned, grouping pixels into coherent regions to capture context relations between distant regions. Based on the captured relation, a graph is generated through the Mahalanobis distance, thereby enhancing the performance of the image classification task.

\subsubsection{Deep Learning-based Method}
Deep learning-based methods in image-to-graph transformation utilize deep learning models to identify object locations in images and extract relations between objects. While image-to-graph transformation using deep learning is not well-established, a few studies have been proposed that employ CNN-based models~\cite{gu_2019_i2g, zhou_2020_i2g}. Figure~\ref{F:image_to_graph}~(b-1) and (b-2) show two approaches that utilize CNN-based models for image-to-graph transformation.

Previous studies convert image features extracted from the encoder into sentences in the decoder. The converted text is used for graph transformation. Diverging from previous approaches, Gu et al.~\cite{gu_2019_i2g} generate a graph by transforming an image into an image scene-graph using RCNN and biLSTM and the image scene-graph into a sentence scene-graph, as depicted in Figure~\ref{F:image_to_graph}~(b-1). Consequently, they addressed the image captioning problem, where no direct pair exists between the image and the generated image graph. On the other hand, Zhou et al.~\cite{zhou_2020_i2g} used the VGG-19 model for image-to-graph transformation to enhance low-resolution images into high-resolution images, as shown in Figure~\ref{F:image_to_graph}~(b-2). The VGG-19, a CNN-based model, was employed to extract internal features from the downsampled image and restore it to high resolution. The extracted image features were then organized into a graph through vertex mapping using k-nearest neighboring.

\subsubsection{Dataset}
The image-to-graph transformation field has seen fewer studies than other data transformation techniques, resulting in limited datasets available for research. Among the datasets used in image-to-graph, the HyperSpectral Image~(HSI) dataset is notable, as shown in Table~\ref{table:i2g_hsdata}. This dataset contains images with diverse land covers, and various properties such as spatial resolution, spectral channels, wavelength range, and land-cover classes vary depending on the specific data. Additionally, Table~\ref{table:i2g_otherdata} presents other image datasets besides the HSI dataset categorized into indoor and outdoor scenes. Note that the column labeled \#\textbf{per Image} indicates the number of reference images corresponding to each query in the dataset.

\begin{table}[!htbp]
    \small
    \caption{HSI datasets used for image-to-graph transformation}
    \centering
    \label{table:i2g_hsdata}
    \begin{tabular}    {p{0.13\textwidth}p{0.05\textwidth}p{0.1\textwidth}p{0.06\textwidth}p{0.1\textwidth}p{0.1\textwidth}p{0.1\textwidth}p{0.05\textwidth}}
        \toprule
            \multicolumn{1}{c}{\textbf{Dataset}} & \multicolumn{1}{c}{\textbf{Pixels}}
            & \begin{tabular}[c]{@{}c@{}}\textbf{Spatial}\\\textbf{resolution}\end{tabular}
            & \begin{tabular}[c]{@{}c@{}}\textbf{Spectral}\\\textbf{channels}\end{tabular}
            & \begin{tabular}[c]{@{}c@{}}\textbf{Wavelength}\\\textbf{range}\end{tabular}
            & \begin{tabular}[c]{@{}c@{}}\textbf{Land-cover}\\\textbf{Class}\end{tabular}
            & \multicolumn{1}{c}{\textbf{Sensors}} & \multicolumn{1}{c}{\textbf{Ref}} \\
        \midrule
            Indian Pines & \multicolumn{1}{c}{145x145} & \multicolumn{1}{c}{20mx20m} & \multicolumn{1}{r}{220} & \multicolumn{1}{c}{0.4$\sim$2.5$\mu$m} & \multicolumn{1}{r}{16} & \multicolumn{1}{c}{AVIRIS} & \multicolumn{1}{c}{\cite{Indian_Pines}} \\
            Pavia University & \multicolumn{1}{c}{610x340} & \multicolumn{1}{c}{1.3mx1.3m} & \multicolumn{1}{r}{103} & \multicolumn{1}{c}{0.43$\sim$0.86$\mu$m} & \multicolumn{1}{r}{9} & \multicolumn{1}{c}{ROSIS} & \multicolumn{1}{c}{-} \\
            Houston 2013 & \multicolumn{1}{c}{349x1,905} & \multicolumn{1}{c}{2.5m} & \multicolumn{1}{r}{144} & \multicolumn{1}{c}{380$\sim$1050$n$m} & \multicolumn{1}{r}{15} & \multicolumn{1}{c}{ITRES CASI-1500}  & \multicolumn{1}{c}{-} \\
            Salinas & \multicolumn{1}{c}{512x217} & \multicolumn{1}{c}{3.7m} & \multicolumn{1}{r}{204} & \multicolumn{1}{c}{-} & \multicolumn{1}{r}{16} & \multicolumn{1}{c}{AVIRIS} & \multicolumn{1}{c}{-} \\
        \bottomrule
    \end{tabular}
\end{table}

\begin{table}[!htbp]
    \small
    \caption{\textcolor{black}{Other image datasets used for image-to-graph transformation}}
    \centering
    \label{table:i2g_otherdata}
    \begin{tabular}{p{0.1\textwidth}p{0.05\textwidth}p{0.05\textwidth}p{0.05\textwidth}p{0.1\textwidth}p{0.15\textwidth}p{0.15\textwidth}p{0.1\textwidth}p{0.05\textwidth}}
        \toprule
            \multicolumn{1}{c}{\textbf{Dataset}} & \multicolumn{1}{c}{\textbf{Year}} & \multicolumn{1}{c}{\textbf{\#Image}} & \multicolumn{1}{c}{\textbf{\#Train}} & \multicolumn{1}{c}{\textbf{\#Test}} & \multicolumn{1}{c}{\textbf{\#Indoor Query}} & \multicolumn{1}{c}{\textbf{\#Outdoor Query}} & \multicolumn{1}{c}{\textbf{\#Per Image}} & \multicolumn{1}{c}{\textbf{Ref}} \\
        \midrule
            rPascal     & \multicolumn{1}{c}{2009} & \multicolumn{1}{r}{4,340} & \multicolumn{1}{r}{2,113} & \multicolumn{1}{r}{2,227} & \multicolumn{1}{r}{18} & \multicolumn{1}{r}{32} & \multicolumn{1}{r}{180} & \multicolumn{1}{c}{\cite{rPascal_dataset}} \\
            rImageNet   & \multicolumn{1}{c}{2009} & \multicolumn{1}{r}{20,121} & \multicolumn{1}{c}{-} & \multicolumn{1}{c}{-} & \multicolumn{1}{r}{14} & \multicolumn{1}{r}{36} & \multicolumn{1}{r}{305} & \multicolumn{1}{c}{\cite{rImaegNet_dataset}} \\
            BSD         & \multicolumn{1}{c}{2001} & \multicolumn{1}{r}{300} & \multicolumn{1}{r}{200} & \multicolumn{1}{r}{100} & \multicolumn{1}{c}{-} & \multicolumn{1}{c}{-} & \multicolumn{1}{c}{-} & \multicolumn{1}{c}{\cite{bsd_dataset}} \\
            DIV2K       & \multicolumn{1}{c}{2017} & \multicolumn{1}{r}{1,000} & \multicolumn{1}{r}{800} & \multicolumn{1}{r}{200} & \multicolumn{1}{c}{-} & \multicolumn{1}{c}{-} & \multicolumn{1}{c}{-} & \multicolumn{1}{c}{\cite{div2k_dataset}} \\
            VG          & \multicolumn{1}{c}{2017} & \multicolumn{1}{r}{101,174} & \multicolumn{1}{c}{-} & \multicolumn{1}{c}{-} & \multicolumn{2}{c}{1,700,000} & \multicolumn{1}{r}{117} & \multicolumn{1}{c}{\cite{VG_dataset}} \\
            MS-COCO     & \multicolumn{1}{c}{2014} & \multicolumn{1}{r}{163,791} & \multicolumn{1}{c}{-} & \multicolumn{1}{c}{-} & \multicolumn{1}{c}{-} & \multicolumn{1}{c}{-} & \multicolumn{1}{r}{5} & \multicolumn{1}{c}{\cite{lin2014microsoft}}   \\
        \bottomrule
    \end{tabular}
\end{table}

\subsubsection{Discussion and Challenges}
\label{dis:img2graph}
The image-to-graph transformation field has seen fewer approaches and studies than other data transformation techniques. This section introduces two methods: one that generates a graph based on the inter-pixel distance for image-to-graph transformation and another that utilizes a deep learning-based model. However, the current research on image-to-graph remains limited in scope, as the transformed graph is mainly used as input data or training datasets for graph-based deep learning models. This narrow focus sets image-to-graph apart from the more diverse and expansive landscape of text-to-graph transformation techniques, demonstrating various examples of transforming text into human-understandable structural graphs, such as knowledge graphs. Expanding research in graph generation, including knowledge incorporation, is crucial as a challenge for image-to-graph technology. This expansion should not be limited to simple models but should also involve transforming data into formats that are easy for humans to understand. By embracing a broader approach, image-to-graph transformation can further evolve and demonstrate its potential in various applications.

\subsection{Video-to-Graph Transformation}

\begin{figure*}[b]
    \centering
    \includegraphics[width=.9\textwidth]{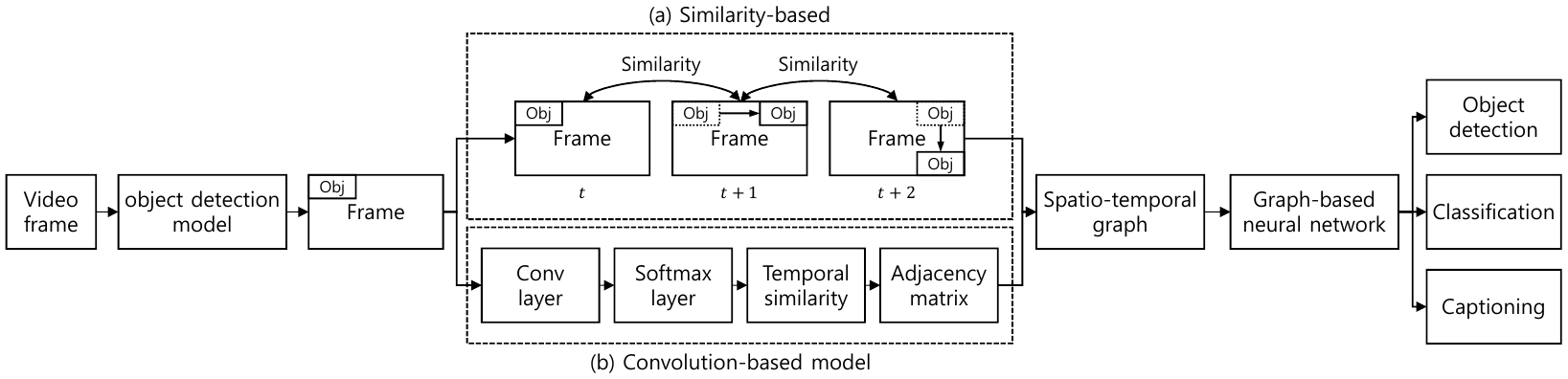}
    \caption{Strategies of video-to-graph. (a) is similarity-based method and (b) is convolution-based model method.
    }
    \label{F:video_to_graph}
\end{figure*}
The video-to-graph technique constructs a spatiotemporal graph by extracting spatial details from individual video frames and temporal relationships between consecutive frames over time. This technique captures the spatial and temporal intricacies within a video, integrating them into a graph-oriented deep-learning framework. Consequently, researchers in the field of video-to-graph are exploring diverse applications such as object detection, captioning, search, and activity recognition. The strategies for video-to-graph transformation, depicted in Figure~\ref{F:video_to_graph}, encompass similarity-based and convolution-based model approaches, categorized, as shown in Figure~\ref{F:video_to_graph}~(a) and (b).

\subsubsection{Similarity-based Method}
The similarity-based method in video-to-graph transformation focuses on establishing relational similarities between neighboring frames, as illustrated in Figure~\ref{F:video_to_graph} (a). The essential point of the similarity-based method is the algorithm or model used for relation calculation to transform the video into a spatiotemporal graph. Zhao et al.~\cite{zhao_2021_v2g_stg} employ a sliding window to compute the similarity between objects in neighboring frames, which leads to the extraction of topological priors. These priors are then utilized for modeling relations between neighboring frames. A pairwise similarity function is also applied for object detection within video frames. Zhang et al.~\cite{zhang_2021_v2g_hop} adopt a multi-hop structure for video clips to calculate the hop-connective relations. They use attention scores to compute similarities, transforming visual features into an interaction latent space to create the graph. This transformed graph is further employed in video self-supervised learning. Soldan et al.~\cite{soldan_2021_v2g} use K-Nearest Neighbors~(K-NN) for temporal relation modeling and establish spatial relation connections between consecutive frames. They apply 1D positional encoding and 1D convolution for dimension mapping of input video data. Pan et al.~\cite{pan_2020_v2g} utilize Intersection over Union~(IoU) for spatial relation modeling and pairwise cosine feature similarities for temporal relation modeling. Their proposed model combines these two relations into a single graph, enhancing the performance of video captioning.

\subsubsection{Convolution-based Model Method}
The convolution-based model method transforms spatial information and temporal relations of video frames into a spatiotemporal graph using a learning layer that calculates similarity, as depicted in Figure~\ref{F:video_to_graph}~(b). Typically, CNN-based models are employed for extracting features from video frames. Zhang and Peng~\cite{zhang_2019_v2g_bd} use a mask R-CNN~\cite{he2017mask} model in video-to-graph transformation for video captioning. The pre-trained CNN extracts spatial region information of salient objects from the video. Temporal dynamics are obtained by computing the trajectories of objects, their temporal order, and similarities among object regions at each frame. This spatial region information and temporal dynamics are transformed into a bidirectional temporal graph. The resulting graph is used as learning data for generating captions through an object-aware aggregation encoder and an attention mechanism decoder in the Vector of Locally Aggregated Descriptors~(VLAD) model. Zhao et al.~\cite{zhao_2021_v2g_pyramid} focus on video retrieval using key areas of the video frame. They extract feature maps from video frames using a pre-trained CNN. Additionally, each frame in a video connects space and time to form a graph. Graph Convolutional Networks~(GCN) is used for learning spatial-temporal correlation. The feature maps of a video frame are transformed into a pyramid regional sub-graph, while the frame sequence is converted into a regional graph. Arnab et al.~\cite{arnab_2021_v2g} aim for video understanding, including reasoning about the relationships between actors, objects, and the environment over the long term. They extract actors, objects, and environment information from video frames using a 3D CNN model. The video is structured to represent the relations and interactions between actors, objects, and environments. These are then transformed into graphs and input into their proposed message-passing graph neural network. Zhang et al.~\cite{zhang_2021_v2g_act} transform video-to-graph for activity recognition. They use the ConvNets model to extract feature sequences from video segments. A single-layer feedforward neural network, consisting of a 1D convolution layer and a softmax layer, is adopted to extract temporal similarity. The adjacency matrix method generates the spatiotemporal graph, which is then input to the GCN model for activity recognition.

\subsubsection{Dataset}
Video-to-graph transformation primarily focuses on utilizing a graph-based deep learning model to address specific tasks rather than merely transforming a video into a graph format. Table~\ref{table:v2g_dataset} presents the datasets employed in video-to-graph transformation. These datasets, generated through video-to-graph transformation, serve as inputs for models designed for various tasks, including object segmentation, action recognition, human action, motion recognition, and video captioning and searching.

\begin{table}[!htbp]
    \small
    \caption{Dataset used for video-to-graph transformation}
    \centering
    \label{table:v2g_dataset}
    \begin{tabular}{p{0.15\textwidth}p{0.05\textwidth}p{0.3\textwidth}p{0.15\textwidth}p{0.15\textwidth}p{0.15\textwidth}p{0.05\textwidth}}
        \toprule
            \multicolumn{1}{c}{\textbf{Dataset}} & \multicolumn{1}{c}{\textbf{Year}} & \multicolumn{1}{c}{\textbf{Category}} & \multicolumn{1}{c}{\textbf{\#Video}} & \multicolumn{1}{c}{\textbf{\#Clip}} & \multicolumn{1}{c}{\textbf{\#Sentence}} & \multicolumn{1}{c}{\textbf{Ref}} \\
        \midrule
            HMDB51       & \multicolumn{1}{c}{2011} & Human motion recognition & \multicolumn{1}{r}{3,312} & \multicolumn{1}{r}{6,766} & \multicolumn{1}{c}{-} & \multicolumn{1}{c}{\cite{HMDB}} \\
            MSVD         & \multicolumn{1}{c}{2011} & Video description & \multicolumn{1}{c}{-} & \multicolumn{1}{r}{1,970} & \multicolumn{1}{r}{70,028} & \multicolumn{1}{c}{\cite{chen2011collecting}} \\
            UCF101       & \multicolumn{1}{c}{2012} & Human action classification & \multicolumn{1}{r}{2,500} & \multicolumn{1}{r}{13,320} & \multicolumn{1}{c}{-} & \multicolumn{1}{c}{\cite{UCF101}} \\
            Kinetics 400 & \multicolumn{1}{c}{2017} & Human action classification & \multicolumn{1}{r}{306,245} & \multicolumn{1}{r}{306,245} & \multicolumn{1}{c}{-} & \multicolumn{1}{c}{\cite{kinetics400}} \\
            JHMDB        & \multicolumn{1}{c}{2013} & Action recognition & \multicolumn{1}{r}{960} & \multicolumn{1}{r}{31,838} & \multicolumn{1}{c}{-} & \multicolumn{1}{c}{\cite{JHMDB}} \\
            Charades     & \multicolumn{1}{c}{2016} & Action recognition & \multicolumn{1}{r}{30} & \multicolumn{1}{r}{9,848} & \multicolumn{1}{r}{27,847} & \multicolumn{1}{c}{\cite{sigurdsson2016hollywood}} \\
            SS-V1        & \multicolumn{1}{c}{2017} & Action recognition & \multicolumn{1}{c}{-} & \multicolumn{1}{r}{198,499} & \multicolumn{1}{c}{-} & \multicolumn{1}{c}{\cite{SSv1}} \\
            SS-V2        & \multicolumn{1}{c}{2018} & Action recognition & \multicolumn{1}{c}{-} & \multicolumn{1}{r}{220,847} & \multicolumn{1}{r}{318,572} & \multicolumn{1}{c}{\cite{SSv2}}\\
            AVA          & \multicolumn{1}{c}{2018} & Action recognition & \multicolumn{1}{r}{504,000} & \multicolumn{1}{c}{-} & \multicolumn{1}{c}{-} & \multicolumn{1}{c}{\cite{AVA}}\\
            ster         & \multicolumn{1}{c}{2019} & Action recognition & \multicolumn{1}{r}{148,092} & \multicolumn{1}{r}{5,331,312} & \multicolumn{1}{c}{-} & \multicolumn{1}{c}{\cite{Jester}} \\
            AG           & \multicolumn{1}{c}{2020} & Action recognition & \multicolumn{1}{r}{100,000} & \multicolumn{1}{c}{-} & \multicolumn{1}{c}{-} & \multicolumn{1}{c}{\cite{ActionGenome}}\\
            TACoS        & \multicolumn{1}{c}{2013} & Video captioning & \multicolumn{1}{r}{127} & \multicolumn{1}{r}{7,206} & \multicolumn{1}{r}{18,227} & \multicolumn{1}{c}{\cite{regneri2013grounding}}\\
            MSR-VTT      & \multicolumn{1}{c}{2016} & Video captioning & \multicolumn{1}{r}{7,180} & \multicolumn{1}{r}{10,000} & \multicolumn{1}{r}{200,000}& \multicolumn{1}{c}{\cite{xu2016msr}}\\
            ANet Captions& \multicolumn{1}{c}{2017} & Video captioning & \multicolumn{1}{r}{20,000} & \multicolumn{1}{r}{100,000} & \multicolumn{1}{r}{100,000} & \multicolumn{1}{c}{\cite{krishna2017dense}} \\
            DiDeMo       & \multicolumn{1}{c}{2017} & Video captioning & \multicolumn{1}{r}{10,642} & \multicolumn{1}{r}{40,000} & \multicolumn{1}{r}{40,000} & \multicolumn{1}{c}{\cite{DiDeMo}} \\
            DAVIS 2017   & \multicolumn{1}{c}{2017} & Object segmentation & \multicolumn{1}{r}{150} & \multicolumn{1}{r}{10,459} & \multicolumn{1}{c}{-} & \multicolumn{1}{c}{\cite{davis_2017_dataset}} \\
            VIP          & \multicolumn{1}{c}{2018} & Object segmentation & \multicolumn{1}{r}{404} & \multicolumn{1}{r}{200,000} & \multicolumn{1}{c}{-} & \multicolumn{1}{c}{\cite{VIP}} \\
            SVD          & \multicolumn{1}{c}{2019} & Video searching & \multicolumn{1}{r}{1,206} & \multicolumn{1}{r}{560,807} & \multicolumn{1}{c}{-} & \multicolumn{1}{c}{\cite{SVD}} \\
        \bottomrule
    \end{tabular}
\end{table}

\subsubsection{Discussion and Challenges}
\label{dis:video2graph}
This section explores the video-to-graph transformation, primarily focusing on capturing the temporal relations between frames through video similarity and deep learning-based methods. However, similar to the issue mentioned in the image-to-graph transformation~(see Section~\ref{dis:img2graph}), the task of transforming video data into a graph is currently limited to being used as input for graph-based deep learning models. Consequently, a significant challenge lies in achieving a video-to-graph transformation that goes beyond this narrow scope and can create a knowledge graph from video data, akin to how text data can be transformed into a knowledge graph for better human understanding. This expansion would enable video-to-graph techniques to discover applications in broader contexts and enhance their utility for various tasks.
\section{Discussion and Conclusion}
\label{sec:conclusion}
In this survey, we introduced data transformation strategies to remove format heterogeneity within heterogeneous data. These techniques serve various purposes, such as preprocessing for data integration, preparing training data for models, transforming data to include necessary information, and converting data into formats easily understandable by humans. Despite its crucial role in leveraging heterogeneous data by eliminating heterogeneity like data fusion and cleaning, data transformation has received less attention than other techniques. Thus, there is a lack of comprehensive surveys encompassing a wide range of methods, including various techniques incorporating the growing use of deep learning among researchers. However, given their role as solutions that reduce human effort in transforming input formats for deep learning models and preparing training data, data transformation strategies employing state-of-the-art technology are highly valuable. This survey aims to guide various purposes and cases by summarizing data transformation strategies technically and addressing heterogeneity resulting from format conflicts. We have introduced data transformation strategies based on source and target formats, categorized them into data-to-text and data-to-graph, and provided explanations accordingly.

The main challenge of the data-to-text in Section~\ref{sec:d2t} appears to be generating high-quality and flexible textual content from various data formats, including tables, text, images, and videos. This involves addressing issues such as hallucination where the generated text does not faithfully represent the source data, handling diverse data structures and formats, ensuring the naturalness and variety of generated text, and incorporating external knowledge sources effectively. Another significant challenge is to interpret how the text was generated and achieve interpretability in text generation models. Overall, the main challenge in Section~\ref{sec:d2t} is to improve the versatility and quality of data-to-text transformations across different data formats while maintaining the fidelity and interpretability of the generated text.

The primary challenge of the data-to-graph in Section~\ref{sec:d2g} is to effectively transform different data formats~(text, images, and videos) into graph representations. This involves overcoming domain-specific variations, linguistic diversity, and inconsistent data formats. Additionally, expanding the scope of graph transformation beyond serving as input for deep learning models and creating knowledge graphs from data like video, similar to text-based transformations, presents a significant challenge with the potential to unlock broader applications in artificial intelligence and data understanding.

Moreover, as mentioned in Section~\ref{sec:heterogeneity}, domain conflicts leading to heterogeneity result from multiple factors. Hence, techniques supporting processes such as data, modeling, visualization, and user decision-making within visual analysis systems are required to resolve domain conflicts~\cite{keim2006challenges}. Additionally, investigating techniques to remove domain conflicts requires preliminary research on complementary strategies like data integration, cleaning, and transformation. For these reasons, in this survey we explored data transformation strategies that are currently under-investigated. As a next step of this survey, we identify the need for an overview of strategies to remove domain conflicts, prompting a survey exploring these strategies. Consequently, we plan to investigate strategies and techniques to resolve domain conflicts for data analysis and decision-making.

\bibliographystyle{ACM-Reference-Format}
\bibliography{heterogeneous_survey}

\end{document}